\documentclass[journal]{IEEEtran}

\usepackage[scaled=1.0]{helvet}
\usepackage{times}
\usepackage{graphicx}
\usepackage{subfigure}
\usepackage{parskip}
\usepackage{multirow}
\usepackage{amsmath}
\usepackage{color}
\usepackage[labelfont=bf,textfont=it]{caption}
\usepackage[lined,boxed,commentsnumbered, ruled]{algorithm2e}

\begin{document}
\title{Multi-Scale Single Image Dehazing Using Laplacian and Gaussian Pyramids}
%
%
% author names and IEEE memberships
% note positions of commas and nonbreaking spaces ( ~ ) LaTeX will not break
% a structure at a ~ so this keeps an author's name from being broken across
% two lines.
% use \thanks{} to gain access to the first footnote area
% a separate \thanks must be used for each paragraph as LaTeX2e's \thanks
% was not built to handle multiple paragraphs
%
\author{
Z. G. Li, H. Y. Shu, and C. B. Zheng
\thanks{Zhengguo Li and Haiyan Shu are with SRO department, the Institute for Infocomm Research, Singapore, 138632, (emails: \{ezgli, hshu\}@i2r.a-star.edu.sg). Chaobing Zheng is with the school of Information Science and Engineering, Wuhan University of Science and Technology, Wuhan 430081, China (e-mail: zhengchaobing@wust.edu.cn)}% <-this % stops a space
}

% note the % following the last \IEEEmembership and also \thanks -
% these prevent an unwanted space from occurring between the last author name
% and the end of the author line. i.e., if you had this:
%
% \author{....lastname \thanks{...} \thanks{...} }
%                     ^------------^------------^----Do not want these spaces!
%
% a space would be appended to the last name and could cause every name on that
% line to be shifted left slightly. This is one of those "LaTeX things". For
% instance, "\textbf{A} \textbf{B}" will typeset as "A B" not "AB". To get
% "AB" then you have to do: "\textbf{A}\textbf{B}"
% \thanks is no different in this regard, so shield the last } of each \thanks
% that ends a line with a % and do not let a space in before the next \thanks.
% Spaces after \IEEEmembership other than the last one are OK (and needed) as
% you are supposed to have spaces between the names. For what it is worth,
% this is a minor point as most people would not even notice if the said evil
% space somehow managed to creep in.

% The paper headers
\markboth{}%Journal of \LaTeX\ Class Files}%,~Vol.~6, No.~1, January~2007}%
{Shell \MakeLowercase{\textit{et al.}}: Bare Demo of IEEEtran.cls
for Journals}
% The only time the second header will appear is for the odd numbered pages
% after the title page when using the twoside option.
%
% *** Note that you probably will NOT want to include the author's ***
% *** name in the headers of peer review papers.                   ***
% You can use \ifCLASSOPTIONpeerreview for conditional compilation here if
% you desire.

% If you want to put a publisher's ID mark on the page you can do it like
% this:
%\IEEEpubid{0000--0000/00\$00.00~\copyright~2007 IEEE}
% Remember, if you use this you must call \IEEEpubidadjcol in the second
% column for its text to clear the IEEEpubid mark.

% use for special paper notices
%\IEEEspecialpapernotice{(Invited Paper)}

% make the title area
\maketitle

\begin{abstract}
 Model-based single image dehazing was widely studied due to its extensive applications. {\it Ambiguity between  object radiance and haze} and {\it noise amplification in sky regions} are two inherent problems of model-based single image dehazing. In this paper, a dark direct attenuation prior (DDAP) is  proposed to address the former problem. A novel haze line averaging is proposed to reduce the morphological artifacts caused by the DDAP which enables a weighted guided image filter with a smaller radius to further reduce the morphological artifacts while preserve the fine structure in the image.  A multi-scale dehazing algorithm is then proposed to address the latter problem by adopting Laplacian and Gaussian pyramids to decompose the hazy image into different levels and applying different haze removal and noise reduction approaches to restore the scene radiance at the different levels. The resultant pyramid is collapsed to restore a haze-free image.  Experiment results demonstrate that the proposed algorithm outperforms state-of-the-art dehazing algorithms.
\end{abstract}


% For peer review papers, you can put extra information on the cover
% page as needed:
% \begin{center} \bfseries EDICS Category: 3-BBND \end{center}
%
% for peerreview papers, inserts a page break and creates the second title.
% Will be ignored for other modes.
\IEEEpeerreviewmaketitle

\section{Introduction}
\label{section1}
Under bad weather conditions such as haze, smoke, fog, and so on, a real-world outdoor hazy image suffers from contrast loss of captured objects \cite{1nara2003} and color distortion \cite{1ancuti2013,1gald2018}. Single image dehazing is to restore a haze-free image from a single hazy image. The color distortion caused by the air-light is corrected as well as both the local and the global contrast of the scene are increased in the restored image \cite{1nara2003}.  The problem was well studied due to its possibly board applications such as objection detection, and so on.

Existing single image dehazing algorithms can be classified into fusion-based, data-driven, and model-based algorithms. It was observed by the fusion-based algorithms \cite{1ancuti2013,1gald2018,1Galdran2017} that a hazy image suffers from contrast loss and color distortion. A multi-scale fusion method was proposed in \cite{1ancuti2013} for single image dehazing. Two enhanced images of a hazy image are first generated. Inspired by the multi-scale exposure fusion algorithm in \cite{1mertens2009}, three weight maps, i.e., luminance map, chromatic map and saliency map, are computed for the two images. The enhanced images  are fused by using the the Gaussian and Laplacian pyramids \cite{1burt1983} to synthesize a restored image. Multiple  over-exposed images are first produced from the hazy image, and then fused via a Laplacian pyramid decomposition method in \cite{1gald2018}. The weights and the images to be fused are learned using variational method in \cite{1Galdran2017}. Unfortunately, all the derived images cannot reflect the depth information of the scene well, resulting in a poor dehazed image in presence of heavy haze \cite{1liuw2021}. In addition, fine details are not preserved well by the multi-scale fusion \cite{LiZG2017}.

There are many data-driven dehazing algorithms, especially deep learning based dehazing algorithms.  Ren et al. \cite{Ren} and Cai et al. \cite{cai}  proposed to study single image dehazing using deep convolutional neural networks (CNNs). All the CNN based algorithms have potential to obtain higher SSIM or PSNR values but the dehazed images look blurry and they are not photo realistic. Their performance also needs to be improved if the haze is heavy \cite{glots2020}. Qu et al. \cite{Qu_2019_CVPR} proposed a generative adversarial nets (GANs) based dehazing algorithm, which generated a haze-free image without using the physical scattering model  \cite{koschmider}. Unlike the CNN based algorithms, the GAN based dehazing algorithm is able to produce photo-realistic images even though generated texture is different from real one. Color distortion is also a possible issue for the deep learning based algorithms. Even though the data-driven dehazing was popular \cite{1qin2020,1dongj2020,1hong2020,1dong2020}, recent investigations indicate that the data-driven algorithms perform well for synthetic images but poor for real-world hazy images, especially for those images with heavy haze \cite{1liuw2021,glots2020}.

Many model-based dehazing algorithms are proposed by using the Koschmieders law \cite{koschmider}. An outdoor hazy image can be modelled by a pixel-wise convex combination of a haze-free image and a global atmospheric light. Since the number of freedoms is larger than the number of observations, single image dehazing is an ill-posed problem. {\it Ambiguity between  object radiance and haze} is one inherent problem for the model-based singe image dehazing algorithms. Different priors were proposed to address the inherent problem \cite{He,CVPR16}. He et al. \cite{He} introduced a well known dark channel prior (DCP) for single image dehazing.  The DCP assumes that  each local patch of an outdoor haze-free image includes at least one  pixel which has at least one color channel with a small intensity. The patch is usually selected as $15\times 15$ by the existing algorithms. The DCP introduces morphological artifacts to the restored image. Guided image filter (GIF) \cite{1he2013} and  weighted GIF (WGIF) \cite{Li2014} can be selected to reduce the morphological artifacts. The radius of the filter is usually chosen as 60 such that the morphological artifacts can be smoothed. On the other hand, fine structure cannot be preserved well by the GIF or WGIF with the large radius \cite{Li2018}.  A novel haze line prior (HLP) was proposed in \cite{CVPR16} to address this  inherent problem. The HLP is based on an observation that the colors of a haze-free image can usually be represented by a finite number of different colors  which are classified into clusters in the RGB space \cite{1orchard1991}. All pixels in each cluster form a haze line in the hazy image. The HLP assumes that each haze line includes at least one haze-free pixel. An initial transmission map  is estimated by using the HLP and the artifacts of the initial transmission map are reduced by using the weighted least squares (WLS) framework \cite{1lag1988,1farb2008,1min2014}. The  priors in \cite{He,CVPR16} are robust to different degree of haze in hazy images. Unfortunately, both the DCP in \cite{He} and the HLP in \cite{CVPR16} are not true for those pixels in the sky region.

The transmission map is very small for a pixel in the sky region due to its large depth. A hazy image can be noisy \cite{1fang2014,1liu2017}. Noise is amplified in the sky region of the dehazed image \cite{Li2014,Li2015}. This implies that {\it noise amplification in the sky region} is the other inherent problem for the Koschmieders law based single dehazing algorithms.  Li et al \cite{Li2014,Li2015} introduced an adaptive sky compensation term to address the problem. However, a pixel in the brightest object could be classified as a pixel in the sky region. Subsequently, haze of the brightest object might not be removed well. Clearly, the latter inherent problem is not well addressed by the existing model-based dehazing algorithms. This could be one reason that mean opinion score values of hazy images are higher than those of dehazed images by the model-based dehazing algorithms \cite{1liuw2021}. It is thus desired to develop a robust single image dehazing algorithm to address both the inherent problems.

In this paper, a novel multi-scale model-based dehazing algorithm is proposed to address  these two inherent problems by using the Laplacian pyramids \cite{1burt1983}. Same as the DCP based dehazing algorithms \cite{He,Li2015}, the transmission map is assumed to be constant in a small neighborhood of each pixel. A new multi-scale hazy image model is first built up. The hazy image is decomposed into multiple scales via the Laplacian pyramid. The transmission map is decomposed into multiple scales by using the Gaussian pyramid. It is found that {\it the Gaussian pyramids of the hazy and haze-free images follow the Koschmieders law at the different levels} \cite{Li2021}. The relationship between their Laplacian pyramids at the different level is derived from the Koschmieders law and the new finding.

A simple multi-scale dehazing algorithm is then proposed on top of the multi-scale hazy image model.  The global atmospheric light is estimated  by using the hierarchical searching method in \cite{Three}. The transmission map is estimated by using a  new dark direct attenuation prior (DDAP). As indicated in \cite{He}, the DCP is not true for those pixels in sky region. The proposed DDAP is extended from the DCP, and it is applicable to all the pixels in the hazy image. Same as \cite{He,Li2014,Li2018}, an initial transmission map is obtained by using the DDAP.  Instead of directly using the edge-preserving smoothing filters as in \cite{He,Li2014,Li2018}, the morphological artifacts caused by the DDAP are first reduced by a novel haze line averaging algorithm. The proposed haze line averaging algorithm first uses the initial transmission map to estimate the distance of each pixel to the origin in the haze free image, and then averages over a set of pixels in the same haze line to obtain an improved estimation on the distance of each pixel to the origin in the haze free image. The improved estimation is applied to compute the transmission map. The proposed haze line averaging well utilizes the grouping manner provided by the haze line, and it is efficient to reduce the morphological artifacts caused by the DDAP.  The WGIF with a small radius is applied to further refine the transmission map, and it can preserve the fine structure in the image as well. Clearly, the proposed algorithm is fundamentally different from the algorithm in \cite{CVPR16} in the sense that the distance of each pixel to the origin in the haze free image is determined by the haze free pixel in \cite{CVPR16} while it is computed by the haze line averaging in the proposed algorithm.

 Since the Laplacian pyramid of the hazy image contains different amount of noise at different level, different noise reduction method is adopted to recover the scene radiance at the different level of the pyramid. The resultant pyramid is finally collapsed to restore a dehazed image. Experimental results show that the proposed algorithm outperforms existing dehazing algorithms from both subjective quality and objective quality (such as dehazing quality index (DHQI) in \cite{1min2019}) points of view, and it can prevent noise from being amplified in the sky region regardless of haze degree.  Overall, three major contributions of this paper are:  1)  a new DDAP  which is applicable to all pixels in a hazy image including those pixels in the sky region; 2) a novel haze line averaging algorithm which can reduce the morphological artifacts caused by the DDAP; and 3) a simple multi-scale dehazing algorithm which can prevent noise from being amplified in the sky region.

The rest of this paper is organized as below. Two inherent problems of model-based single image dehazing are analyzed in Section \ref{analysis}. A new Laplacian pyramid hazy image model is introduced in Section \ref{model}. A novel multi-scale dehazing algorithm is proposed in Section \ref{theory} to address both the inherent problems.  Experimental results are provided  to verify the proposed algorithm in Section \ref{experiment}.  Finally, conclusion remarks are  provided  in Section \ref{conclusion}.

\section{Inherent Problems of Model-Based DeHazing}
\label{analysis}

Suppose that $Z$ is a hazy image and $I$ is the corresponding haze-free image. The relationship between them can be represented by the following Koschmieders law \cite{koschmider}:
\begin{equation}
\label{eq1}
Z_c(i,j) = I_c(i,j)t(i,j) + A_c(1-t(i,j)),
\end{equation}
where $(i,j)$ is a pixel position, and $c\in\{R, G, B\}$ is a color channel.  $A_c$ is the atmospheric light and is a constant for the whole image.  $t$ is the transmission map which describes the portion of the light reaching the camera.

The term $I_c(i,j)t_c(i,j)$ is called direct attenuation, and the term $A_c(1-t(i,j))$ is called airlight. When the atmosphere is  homogenous, $t(i,j)$ can be computed by $\exp^{-\alpha d(i,j)}$, $\alpha(>0)$ represents the scattering coefficient of the atmosphere. $d(i,j)$ is the depth of pixel $(i,j)$.

Both the atmospheric light $A$ and the transmission map $t$ are  estimated to restore a haze-free image. Consider a pixel $(i,j)$ in the sky region, the transmission map $t(i,j)$ is almost zero and the value of $Z_c(i,j)$ is $A_c$. Thus, the variance of pixel values is usually very small and the average of pixel values is usually large in the sky region. The atmospheric light $A_c$ can be estimated as the brightest color in the hazy image.  A hierarchical searching method was proposed in \cite{Three} to estimate the atmospheric light $A$ by using these observations.

\subsection{Ambiguity between  Object Radiance and Haze}

It is challenging to estimate the transmission map $t(i,j)$ for each pixel. The {\it ambiguity between  object radiance and haze} is one inherent problem for the model-based singe image dehazing because there are only three equations but at least four unknown variable per pixel position.  The challenging problem can be resolved by introducing different priors  \cite{He,1zhuq2015,CVPR16,1liu2021}. Two most relevant priors, DCP and HLP, are detailed as below.

{\it 1) The DCP}: The DCP in \cite{He} is the most popular prior. The dark channel of an image $I(i,j)$ is defined as
\begin{align}
f_{\rho}(I)(i,j)=\min_{(i',j')\in \Omega_{\rho}(i,j)}\{\min_{c\in\{R, G, B\}}\{I_c(i',j')\}\},
\end{align}
where $\Omega_{\rho}(i,j)$ is a square window centered at the pixel $(i,j)$ of a radius $\rho$ which is usually selected as 7 \cite{He}.

By assuming that  $t(i,j)$ is constant in a neighborhood of $(i,j)$, it can be derived from the equation (\ref{eq1}) that
\begin{align}
\label{fdza}
f_{\rho}(\frac{Z}{A})(i,j)\approx f_{\rho}(\frac{I}{A})(i,j)+1-t(i,j).
\end{align}
The DCP in \cite{He} assumes that $f_{\rho}(I)(i,j)$ is zero for each pixel $(i,j)$. It can be derived from the DCP that
\begin{align}
\label{tijrho}
t(i,j)\approx 1-f_{\rho}(\frac{Z}{A})(i,j).
\end{align}

There are visibly  morphological artifacts if the $t(i,j)$ is directly applied to remove the haze from a hazy image. An edge preserving smoothing filter is usually adopted to refine the transmission map $t(i,j)$. However, fine structure might not be well preserved if the GIF in \cite{1he2013} or the WGIF in \cite{Li2014} is selected due to its large radius \cite{Li2018}. Even though the problem can be addressed by the globally GIF (GGIF) in \cite{Li2018}, the complexity of the GGIF is an issue.

{\it 2) The HLP}: Define a set $H(i'',j'')$ as
\begin{align}
\label{hijij}
\{(i,j)|\sum_{c\in \{R, G,B\}}|I_c(i'',j'')-I_c(i,j)|=0\}.
\end{align}

All the $Z(i,j)$'s with $(i,j)$ in the set $H(i'',j'')$ form a haze line with  the color $I(i'',j'')$ and the atmospheric light $A$  as the two end points \cite{CVPR16}. Each haze line can be identified by using a  color shift hazy pixel $\hat{Z}(i,j)$ which is defined as $(Z(i,j)-A)$. The $\hat{Z}(i,j)$ can be converted into the spherical coordinates, $[r(i,j)$, $\theta(i,j)$, $\psi(i,j)]$ as
\begin{align}
\left\{\begin{array}{l}
\hat{Z}_R(i,j) = r(i,j)\sin\psi(i,j)\cos\theta(i,j)\\
\hat{Z}_G(i,j) = r(i,j)\sin\psi(i,j)\sin\theta(i,j)\\
\hat{Z}_B(i,j) = r(i,j)\cos\psi(i,j)
\end{array}
\right.,
\end{align}
where $\theta(i,j)$ and $\psi(i,j)$ are the longitude and latitude, respectively, and  $r(i,j)$ is
\begin{align}
\label{rriijj}
r(i,j)=\|\hat{Z}(i,j)\|=t(i,j)\|I(i,j)-A\|.
\end{align}
Consider two pixel positions $(i,j)$ and $(i',j')$ in the set $H(i'',j'')$, it can be verified that
\begin{align}
\theta(i,j)=\theta(i',j'), \psi(i,j)=\psi(i',j').
\end{align}
In other words, all pixels are clustered into different haze lines according to $[\theta(i,j)$, $\psi(i,j)]$.
The pixels in one cluster are usually located at different distances from the camera.  It was shown in \cite{1orchard1991} that the colors of a haze-free image can usually be represented by several hundred different colors. Thus, all the $Z(i,j)$'s form several hundred haze lines.

The HLP  in \cite{CVPR16} assumes that  each haze line includes a haze-free pixel. The HLP implies that
\begin{align}
r_{max}(i'',j'')=\|I(i,j)-A\|\; ;\; \forall (i,j)\in H(i'',j''),
\end{align}
where $r_{max}(i'',j'')$ is defined as
\begin{align}
\label{maximum}
r_{max}(i'',j'')=\max_{(i,j) \in H(i'',j'')}\{r(i,j)\}.
\end{align}

With the HLP, an initial  transmission map $t(i,j)$ can be estimated as
\begin{align}
\label{transmissionmap}
t(i,j)=\frac{1}{r_{max}(i'',j'')}r(i,j)
\end{align}
for any pixel $(i,j)$ in the set $H(i'',j'')$.  A lower bound is then defined for the $t(i,j)$ as
\begin{align}
\varsigma(i,j)=1-f_0(\frac{Z}{A})(i,j),
\end{align}
and $t(i,j)$ is refined as $\max\{t(i,j), \varsigma(i,j)\}$.

The transmission map is finally refined by solving minimizing the following cost function \cite{CVPR16}:
\begin{align}
\label{wlssmooth}
\sum_{(i,j)}[\frac{(t^*(i,j)-t(i,j))^2}{\sigma^2(i,j)}+\lambda \sum_{(i',j')\in N_{i,j}}\frac{(t^*(i,j)-t^*(i',j'))^2}{\|Z(i,j)-Z(i',j')\|^2}],
\end{align}
where $\lambda$ is a parameter which controls trade-off between two terms, $N_{i,j}$ denotes the four nearest
neighbors of $(i,j)$ and $\sigma(i,j)$ is the standard deviation of $t(i,j)$ in each cluster \cite{CVPR16}. The resultant global edge-preserving smoothing filter also belongs to the WLS framework \cite{1lag1988,1farb2008,1min2014}.  Same as the GGIF in \cite{Li2018}, the complexity of minimizing the cost function  (\ref{wlssmooth}) is an issue even though a simpler method was proposed in \cite{1min2014}.

 \subsection{Noise Amplification in  Sky Region}

 Once the atmospheric light $A$ and the transmission map $t^*$ are available,  the haze-free image $I$ can be recovered by:
\begin{equation}
\label{eq2}
I(i,j) = \frac{{Z(i,j)-A}}{\max\{t^*(i,j),{t_L}\}}+A,
\end{equation}
where $t_L$ is a small constant such as 0.1 in  \cite{He}.

Most existing haze removal algorithms assume that the hazy image $Z$ is noise free. However, an outdoor hazy image $Z$ is usually noisy \cite{Li2014,1fang2014,1liu2017}, and it can be represented by
\begin{equation}
\label{eq31}
Z(i,j) = \tilde{Z}(i,j) +  n(i,j),
\end{equation}
where  $\tilde{Z}$ is a hazy but noise free image and $n$ is the noise image. Combining the equations (\ref{eq2}) and (\ref{eq31}), the haze-free image is restored by
\begin{equation}
\label{eq4}
I(i,j) = \frac{\tilde{Z}(i,j)-A}{\max\{t^*(i,j),t_L\}}+A + \frac{n(i,j)}{\max\{t^*(i,j),t_L\}}.
\end{equation}

Considering a pixel position $(i,j)$ in the sky region,  $d(i,j)$ is very large,  and $t^*(i,j)$ is almost zero.  Subsequently, $\max\{t^*(i,j),t_L\}$ is equal to $t_L$. The noise is amplified as $\frac{n(i,j)}{t_L}$, as shown in the equation (\ref{eq4}). This implies that {\it noise amplification in the sky region} is the other inherent problem for the model-based single image dehazing. Actually, the atmospheric light $A$ and the transmission map $t^*(i,j)$ could also be affected by the noise if they are estimated from the noisy hazy image $Z$.

Unlike the former inherent problem, the latter inherent problem is not well addressed for the model-based haze removal algorithms.  Li et al \cite{Li2014,Li2015} introduced an adaptive sky compensation term to address the problem. However, pixels in the brightest object could be classified as those in the sky region. Subsequently, haze of the brightest object might not be removed well.

\section {A Laplacian Pyramid Hazy Image Model}

\label{model}
Image decomposition on top of the Laplacian pyramid \cite{1mertens2009,1burt1983} is much simpler than that based on edge-preserving smoothing filters \cite{1he2013,Li2014,1min2014}. In this section, a unique multi-scale hazy image model is built up  by using the  Koschmieders law (\ref{eq1}) and the Laplacian pyramid.

A Laplacian pyramid is usually generated on top of a Gaussian pyramid  \cite{1burt1983}.
 Let $\{Z\}_G^l$ be the $l(0\leq l\leq L_0)$-th level in the Gaussian pyramid of the image $Z$. For simplicity, $L_0$ is chosen as 1 in this paper.  The pyramid is generated as \cite{1burt1983}
\begin{align}
\label{gaussian}
\{Z\}_G^l=\Psi(\{Z\}_G^{l-1})\,
\end{align}
where $\{Z\}_G^0$ is $Z$, $l=1,\cdots,L_0$, and $\Psi(\cdot)$ is defined as
\begin{align}
\label{reduce}
\left\{\begin{array}{l}
\{Z\}_G^l(i,j)={\displaystyle\sum_{m=-1}^{1}\sum_{n=-1}^1}w(m,n)\{Z\}_G^{l-1}(2i+m,2j+n)\\
w(m,n) =\tilde{w}(m)\tilde{w}(n)\\
\tilde{w}(-1)= \tilde{w}(1)= \frac{1}{4}\; ;\;
\tilde{w}(0)=\frac{1}{2}
\end{array}
\right..
\end{align}

  Let $\{Z\}_L^l$ be the  Laplacian pyramid of the image $Z$ and it is generated by \cite{1burt1983}
\begin{align}
\label{laplacian}
\left\{\begin{array}{l}
\{Z\}_L^{l-1}=\{Z\}_G^{l-1}-\{Z\}_E^{l-1}\\
\{Z\}_E^{l-1}=\Phi(\{Z\}_G^l)
\end{array}
\right.,
\end{align}
where $\Phi(\cdot)$ is the reverse of $\Psi(\cdot)$ and it is defined as \cite{1burt1983}
\begin{align}
\{Z\}_E^{l-1}(i,j)=4{\displaystyle\sum_{m=-1}^{1}\sum_{n=-1}^1}w(m,n)\{Z\}_G^{l}(\frac{i-m}{2},\frac{j-n}{2}),
\end{align}
and only terms for which $(i-m)/2$ and $(j-n)/2$ are integers are included in the sum.

 The Gaussian pyramid of  the transmission map $t$ is denoted as $\{t\}_G^l$. Similar to the existing dehazing algorithms \cite{He, Li2015}, the transmission map $t$ is assumed to be constant in a small neighborhood of each pixel. It can be easily shown that
 \begin{align}
 \label{trans}
 \{t\}_G^l(i,j)=\{t\}_G^{l-1}(2i+m,2j+n)=\{t\}_E^{l-1}(2i,2j)
 \end{align}
 hold for all $l(0\leq l\leq L_0)$'s and $m,n\in\{-1,0,1\}$ if $L_0$ is small.

 For simplicity, the hazy image $Z$ is first assumed to be noise free \cite{Li2021}. It can be derived from the equations (\ref{eq1}), (\ref{gaussian}), (\ref{reduce}) and (\ref{trans}) that
 \begin{align}
 \nonumber
 \{Z\}_G^{1}(i,j)&=\{I\}_G^{1}(i,j)t(2i,2j)+A(1-t(2i,2j))\\\label{above}
 &=\{I\}_G^{1}(i,j)\{t\}_G^{1}(i,j)+A(1-\{t\}_G^{1}(i,j)),
 \end{align}
 and from the equations (\ref{laplacian}), (\ref{trans}) and (\ref{above}) that
 \begin{align}
 \nonumber
 \{Z\}_L^{0}(i,j)&=Z(i,j)-\{Z\}_E^{0}(i,j)\\\nonumber
 &=(I(i,j)-\{I\}_E^{0}(i,j))t(i,j)\\\nonumber
 &=\{I\}_L^{0}(i,j)t(i,j)\\
 &=\{I\}_L^{0}(i,j)\{t\}_G^{0}(i,j).
 \end{align}

 Similarly, it can be derived that
 \begin{align}
 \label{newmodel1}
 \left\{\begin{array}{l}
  \{Z\}_G^{l}(i,j)=(\{I\}_G^{l}(i,j)-A)\{t\}_G^{l}(i,j)+A\\
  \{Z\}_L^{l}(i,j)=\{I\}_L^{l}(i,j)\{t\}_G^{l}(i,j)
  \end{array}
  \right.
 \end{align}
 holds for all $l(0\leq l\leq L_0)$'s. Clearly, the Gaussian pyramids of the hazy and haze-free images also follow the Koschmieders law at the different levels. The Laplacian components of the hazy image are scaled down. As a result, the contrast of the hazy image is reduced.

 Similar results can be derived for a noisily hazy image $Z$. The proposed multi-scale hazy image model is finally presented as
 \begin{align}
 \label{newmodel}
 \left\{\begin{array}{l}
  \{Z\}_G^{L_0}(i,j)=(\{I\}_G^{L_0}(i,j)-A)\{t\}_G^{L_0}(i,j)+A\\
  \{Z\}_L^{l}(i,j)=\{I\}_L^{l}(i,j)\{t\}_G^{l}(i,j)+\{n\}_L^l(i,j)
  \end{array}
  \right.,
 \end{align}
 where $0\leq l<L_0$.

 Compared with the single scale hazy image model (\ref{eq1}), the proposed model (\ref{newmodel}) has a distinctive advantage to separate noise as follows:

  1) The low pass component  $\{Z\}_G^{L_0}$ is usually noise free;

  2) The high pass component at the $(l-1)$th layer, $\{Z\}_L^{l-1}$ includes more noise than the high pass component at the $l$th layer  $\{Z\}_L^{l}$. In other words, $\{n\}_L^{l}(i,j)\ll\{n\}^{l-1}_L(i,j)$.

  \begin{figure}[!htb]
\centering{
\subfigure[ Low pass component]{\includegraphics[width=1.63in]{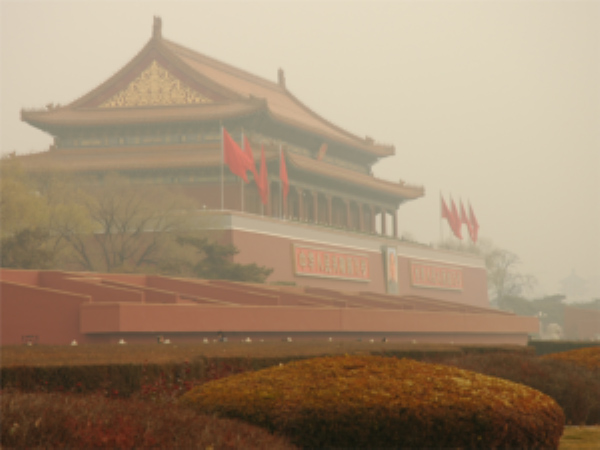}}
\subfigure[ High pass component]{\includegraphics[width=1.63in]{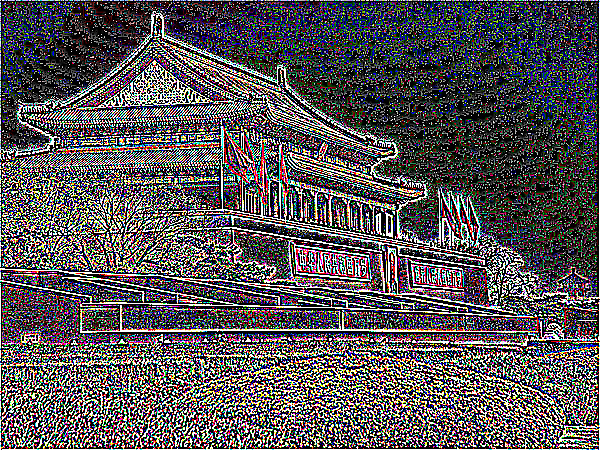}}}
\caption{Low and high pass components of a hazy image.}
\label{Fig3Fig3}
\end{figure}

  One example is shown in Fig. \ref{Fig3Fig3} with $L_0$ as 1. The high pass component is indeed noisy.
  A novel multi-scale single image dehazing algorithm will be designed in the next section by utilizing the above properties.

  \section{Multi-Scale Haze Removal}

\label{theory}

In this section,  the {\it ambiguity between  object radiance and haze} is addressed by extending the DCP \cite{He}, and the  {\it noise amplification in the sky region} is studied by a new multi-scale dehazing algorithm. Since the image $\{Z\}_E^0$ includes less noise than the hazy image $Z$,  both the  atmospheric light $A$ and the transmission map $t$ are estimated by using $\{Z\}_E^0$ rather than the hazy image $Z$ as in \cite{He,CVPR16,Li2015}. The atmospheric light $A$ is estimated by using the method in \cite{Three}.

\subsection{Estimation of the Transmission Map}

The DCP in \cite{He} is not true for those pixels in the sky region \cite{He}. Here, a  new DDAP is provided to replace the DCP. The dark direct attenuation of a hazy image $Z$ is defined as $f_{\rho}(I*t)(i,j)$ which is
\begin{align}
f_{\rho}(I*t)(i,j)=\min_{(i',j')\in \Omega_{\rho}(i,j)}\{\min_{c\in\{R, G, B\}}\{I_c(i',j')t(i',j')\}\}.
\end{align}

The DDAP assumes that $f_{\rho}(I*t)(i,j)$ is zero for the pixel $(i,j)$. Since $t(i,j)$ is zero if the pixel $(i,j)$ is in a sky region, the DDAP is zero for each pixel in the sky region. This implies that the DDAP is less restricted than the DCP in \cite{He}.

The approximation (\ref{tijrho}) can also be derived from the DDAP. In practice, an initial transmission map $t_0(i,j)$ is estimated by
\begin{align}
\label{transmap}
t_0(i,j)=1-\frac{31}{32} f_{\rho}(\frac{Z}{A})(i,j).
\end{align}

\begin{figure*}[!htb]
\centering{
\subfigure[an initial transmission map]{\includegraphics[width=1.63in]{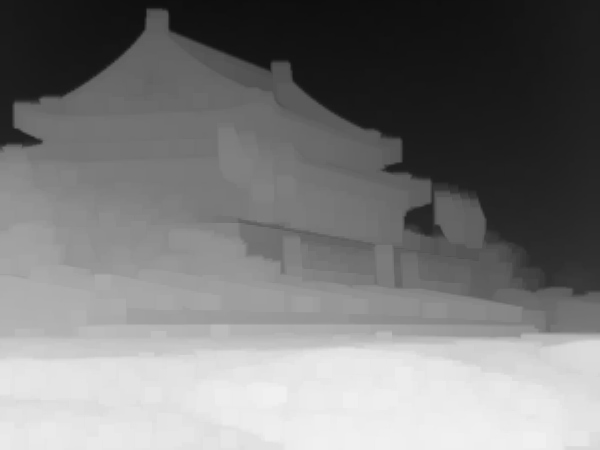}}
\subfigure[a refined transmission map by haze line averaging]{\includegraphics[width=1.63in]{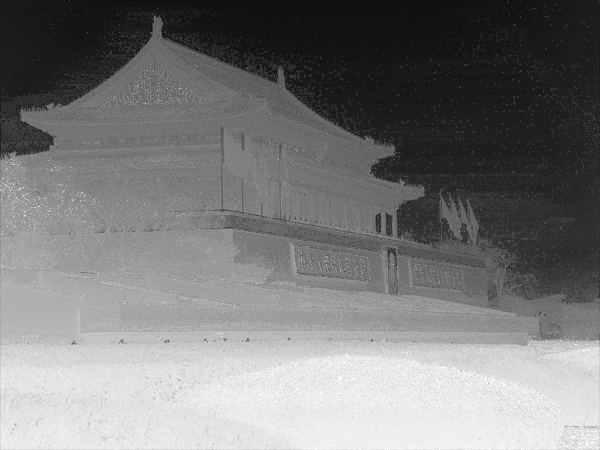}}
\subfigure[a dehazed image by initial transmission map]{\includegraphics[width=1.63in]{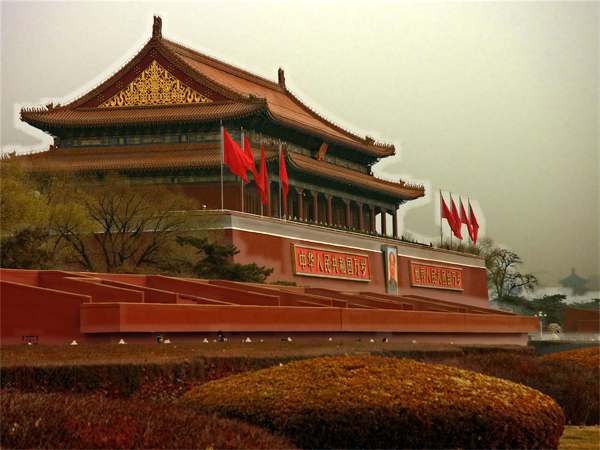}}
\subfigure[a dehazed image by refined transmission map]{\includegraphics[width=1.63in]{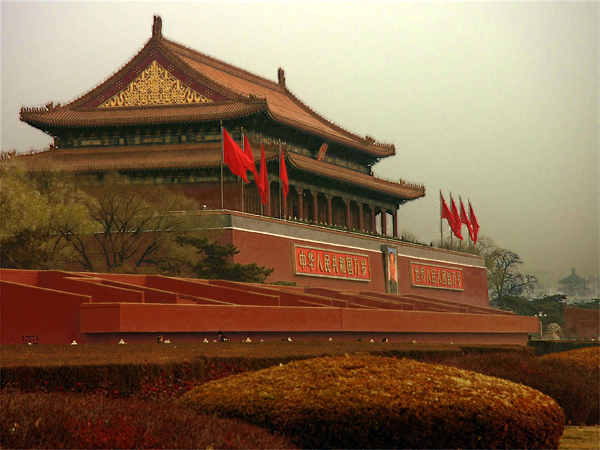}}
}
\caption{Comparison of initial and refined transmission maps as well as their corresponding dehazed images.}
\label{FigFig2}
\end{figure*}

 As shown in Fig. \ref{FigFig2}(c), there are visibly  morphological artifacts if the $t_0(i,j)$ is directly applied to remove the haze from a hazy image. The concept of haze line \cite{CVPR16} is adopted to reduce the morphological artifacts.

It can be derived from the equations (\ref{eq1}) and (\ref{transmap}) that
\begin{align}
\frac{t_0(i,j)}{\|\hat{Z}(i,j)\|}=\frac{1}{\|I(i,j)-A\|}+e(i,j),
\end{align}
where $e(i,j)$ is the morphological artifact caused by the DDAP. Inspired by the nonlocal image processing \cite{1buades2005,1dabov2007}, a set of pixels $(i',j')$'s needs to be found in the hazy image $Z$  such that
\begin{align}
\label{requirement}
\frac{1}{\|I(i',j')-A\|}=\frac{1}{\|I(i,j)-A\|}.
\end{align}
The $1/\|I(i,j)-A\|$ can then be estimated via the following weighted averaging:
\begin{align}
\label{rrmaxij11}
\frac{1}{\|I(i,j)-A\|}=\frac{1}{\mu_{sum}}\sum_{i',j'}\mu(i',j')\frac{t_0(i',j')}{\|\hat{Z}(i',j')\|},
\end{align}
where $\mu(i',j')$ is a weight, and $\mu_{sum}$ is $\sum_{(i',j')}\mu(i',j')$.

All the pixels in the same haze line defined by the equation (\ref{hijij}) happen to meet the requirement (\ref{requirement}). Therefore, the haze line can be adopted to reduce the morphological artifacts. A 2-D histogram binning of $\theta$ and $\psi$ with uniform edges in the range $[0,2\pi]\times [0,\pi]$ is adopted to generate an initial set of $H(i'',j'')$. The bin size is chosen as $\pi/720\times \pi/720$ if not specified in this paper. As indicated in \cite{CVPR16}, this simple method does not generate a uniform sampling of a sphere. An upper bound $\nu$ is then defined for the cardinality of the final sets.  $\nu$ is selected as 200 if not specified. The set $H(i'',j'')$ is dividedly into $\max\{1,|H(i'',j'')|/\nu)$ sub-sets if it is not an empty set. Here, $|H(i'',j'')|$ is the cardinality of  the set $H(i'',j'')$. The sub-sets are denoted as $H_s(i'',j'')$'s ($1\leq s\leq \max\{1,|H(i'',j'')|/\nu)$).

A simple choice of $\mu(i',j')$ is $\|\hat{Z}(i',j')\|$ which happens to perform well. The choice is adopted in this paper. It can derived from the equation (\ref{rrmaxij11}) that
\begin{align}
\label{rrmaxij1166}
\frac{1}{\|I(i,j)-A\|}=\frac{ \sum_{(i',j')\in H_s(i'',j'')}t_0(i',j')}{\sum_{(i',j')\in H_s(i'',j'')}\|\hat{Z}(i',j')\|}.
\end{align}
There are many other different ways to determine the weight $\mu(i',j')$. We leave it to our readers to explore different choices of $\mu(i',j')$. For the pixel $(i,j)$ in the set $H_s(i'',j'')$, it can be computed from the equations (\ref{rriijj}) and (\ref{rrmaxij1166}) that
\begin{align}
\label{myrrmaxij11}
t(i,j)=\frac{ \sum_{(i',j')\in H_s(i'',j'')}t_0(i',j')}{\sum_{(i',j')\in H_s(i'',j'')}\|\hat{Z}(i',j')\|}\|\hat{Z}(i,j)\|.
\end{align}

The  weighted averaging  (\ref{rrmaxij11}) is called haze line averaging, and it is a nonlocal averaging \cite{1buades2005,1dabov2007}. As illustrated in Fig. \ref{FigFig2}(b) and Fig. \ref{FigFig2}(d), it is efficient to reduce the morphological artifacts  caused by the DDAP. It should be noted that the equations (\ref{rrmaxij1166}) and (\ref{myrrmaxij11}) hold if the subset $H_s(i'',j'')$ is replaced by the set $H(i'',j'')$.

The $t(i,j)$  is finally refined by using the WGIF in \cite{Li2014} with the following guidance image \cite{Li2015}:
\begin{align}
\label{guidanceimage}
G(i,j)=1-\min_{c\in \{R, G,B\}}\{\frac{Z_c(i,j)}{A_c}\}.
\end{align}
The   $t(i,j)$  and  $G(i,j)$ are assumed to satisfy
\begin{equation}
\label{linearmodel1} t(i,j)=a_{i',j'}G(i,j)+b_{i',j'}, \forall (i,j)\in
\Omega_{\rho}(i',j').
\end{equation}

The optimal values of $a_{i',j'}$ and $b_{i',j'}$ are derived by minimizing
\begin{eqnarray}
\label{Eap''bp''new}
\sum_{(i,j)\in
\Omega_{\rho}(i',j')}[\Gamma^{G}_{i',j'}(a_{i',j'}G(i,j)+b_{i',j'}-t(i,j))^2+\lambda a_{i',j'}^2],
\end{eqnarray}
 where $\Gamma^{G}_{i',j'}$ is an edge-aware weight. $\rho$ and $\lambda$ are 25 and 1/1000, respectively.  It is worth noting that the value of $\rho$ is 25 rather than 60 as in \cite{Li2014,Li2015}. As shown in \cite{Li2018}, fine structures are preserved better by selecting a smaller $\rho$.

 Once the optimal values of $a_{i',j'}$ and $b_{i',j'}$ are available, the refined transmission map is computed as
 \begin{equation}
 \label{tstar}
t^*(i,j)=\bar{a}_{i,j} G(i,j)+\bar{b}_{i,j},
\end{equation}
where $\bar{a}_{i,j}$ and $\bar{b}_{i,j}$ are the average values of $a_{i',j'}$ and $b_{i',j'}$ in the window $\Omega_{\rho}(i,j)$.

\begin{figure*}[!htb]
\centering{
\subfigure[ by the algorithm in \cite{Li2015}]{\includegraphics[width=1.6in]{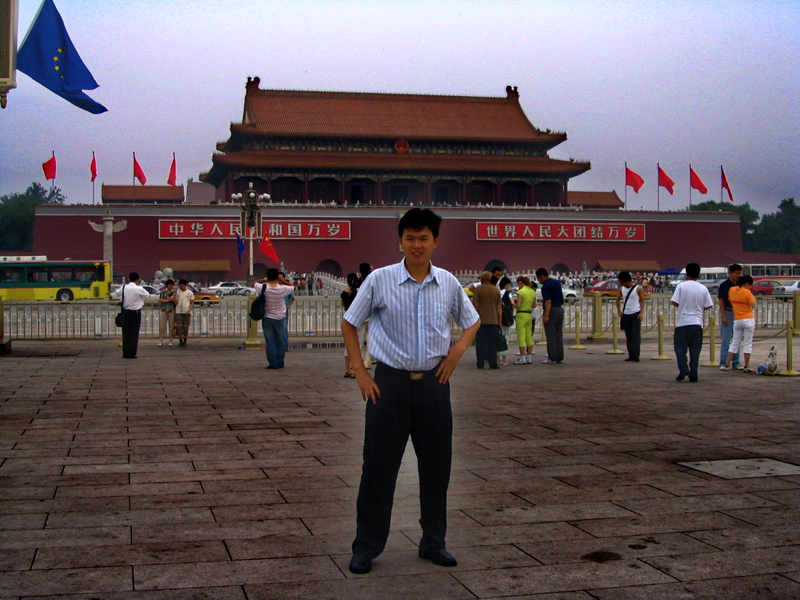}}
\subfigure[ by extending the algorithm in \cite{Li2015} to two scales]{\includegraphics[width=1.6in]{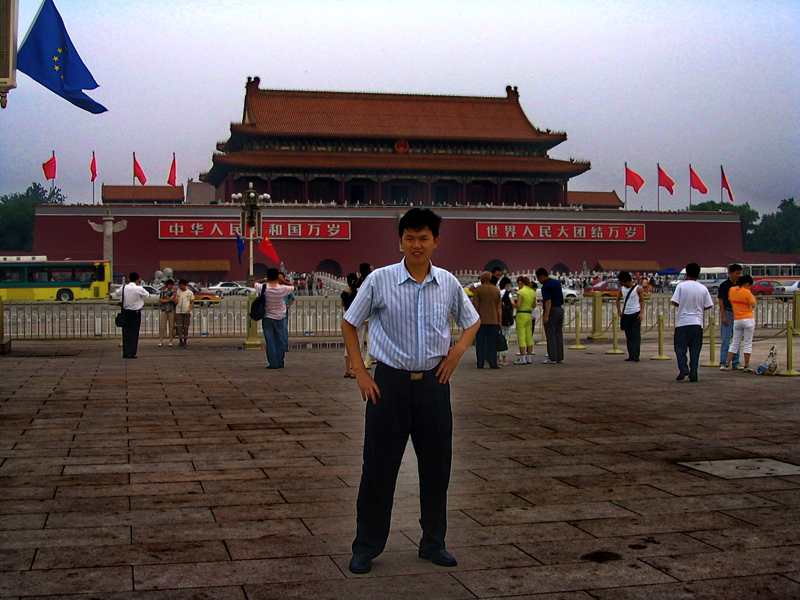}}
\subfigure[ by the proposed algorithm with radius as 25]{\includegraphics[width=1.6in]{pic/Zhengguo_Beijing_2005_Dehaze_I2R_WGIF_4.jpg}}
\subfigure[ by the proposed algorithm with radius as 60]{\includegraphics[width=1.6in]{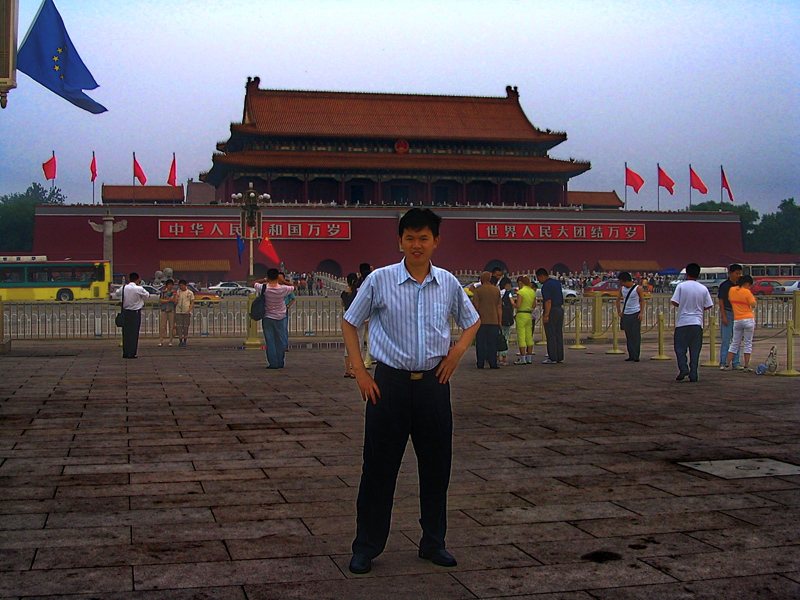}}}
\caption{Four dehazed images by using the algorithm in \cite{Li2015},  extending the algorithm in \cite{Li2015} to two scales, and the proposed algorithm with two different radiuses.}
\label{Fig3}
\end{figure*}

 \subsection{Restoration of The Haze Free Image}
 \label{algorithm}

Since the Laplacian pyramid of the hazy image contains different amount of noise at the  different level, different noise reduction methods are proposed to recover the scene radiance at the different levels of the pyramid.

At the coarsest level $L_0$, the Gaussian component of the scene radiance is restored as
\begin{align}
\label{eta}
\{I\}_G^{L_0}(i,j)=\frac{\{Z\}_G^{L_0}(i,j)-A}{\max\{\{t^*\}_G^{L_0}(i,j),\eta)}+A,
\end{align}
where $\eta$ is a constant. It is adaptive to the haze degree, and should be decreased if the haze degree in increased. It is 1/4 for normal haze and 1/8 for heavy haze in this paper.

At the level $l(0\leq l<L_0)$, the Laplacian component of the scene radiance is recovered by
\begin{align}
\nonumber
\{I\}_L^l(i,j)=&(1-\frac{\phi(\{t^*\}_G^l(i,j))}{2^{l}})\frac{\{Z\}_L^l(i,j)}{\max\{\{t^*\}_G^l(i,j),\eta\}}\\\label{air1}
&+\frac{\phi(\{t^*\}_G^l(i,j))}{2^{l}}\psi(\{t^*\}_G^l(i,j))\{Z\}_L^l(i,j),
\end{align}
where the function of $\phi(\cdot)$ is to define an amplification factor. The $\phi(\cdot)$ and $\psi(\cdot)$ are defined as
\begin{equation}
\label{eq8a}
\left\{\begin{array}{l}
\phi(\{t^*\}_G^l(i,j)) =  \frac{1}{ 1 + \exp(32(\frac{\{t^*\}_G^l(i,j)}{\eta}-1))}\\
\psi(\{t^*\}_G^l(i,j)) =  \frac{\{t^*\}_G^l(i,j)}{\eta}+1
\end{array}
\right..
\end{equation}

The pyramids of $\{I\}_G^{L_0}$ and $\{I\}_L^l(0\leq l<L_0)$ are collapsed to obtain the haze-free image $I$. All the Laplacian components $\{I\}_L^l(i,j)$'s can be tuned according to users' preference. For example, they can be amplified to produce a sharper haze-free image. The global contrast can also be increased. The proposed algorithm is summarized as follows:

\begin{algorithm}
\label{algo3}
{
\caption{Multi-scale model-based single image dehazing}}
\begin{enumerate}
\item[Step 1.] Decompose the image $Z$ into two scales as in the equation (\ref{newmodel}).
\item[Step 2.] Estimate the airlight $A$ from the image $\{Z\}_E^0$.
\item[Step 3.] Estimate the initial transmission map $t_0$ by using the DDAP as in the equation (\ref{transmap}).
\item[Step 4.] Reduce the morphological artifacts of $t_0$ via the nonlocal haze line averaging (\ref{myrrmaxij11}).
\item[Step 5.] Further refine the transmission map using the WGIF as in the equations (\ref{guidanceimage})-(\ref{tstar}).
\item[Step 6.] Restore the haze free components at different levels via the multi-scale dehazing algorithm (\ref{eta})-(\ref{eq8a}).
\item[Step 7.] Collapse  the haze free pyramids to obtain the haze-free image $I$.
\end{enumerate}
\end{algorithm}

\subsection{Theoretical Analysis of the Proposed Algorithm}

Two important case studies of the equation (\ref{air1}) are presented as follows:

{\it Case 1}: $\{t^*\}_G^l(i,j)$ is less than $\eta$, $\phi(\{t^*\}_G^l(i,j))$ is almost 1. It can be derived from the equation (\ref{air1}) that
{\small \begin{align}
\{I\}_L^l(i,j)\approx (1-\frac{1}{2^{l}})\frac{\{Z\}_L^l(i,j)}{\eta}+\frac{1}{2^{l}}\psi(\{t^*\}_G^l(i,j))\{Z\}_L^l(i,j).
\end{align}}
Considering $\{t^*\}_G^l(i,j)\approx 0$ when $l=0$, and $l=1$, it can be derived that
\begin{align}
\left\{\begin{array}{l}
\{I\}_L^1(i,j)\approx \frac{1}{2}\frac{\{Z\}_L^l(i,j)}{\eta}+\frac{1}{2}\{Z\}_L^l(i,j)\\
\{I\}_L^0(i,j)\approx \{Z\}_L^1(i,j)
\end{array}
\right..
\end{align}

Since the noise is reduced from the finest level to the coarsest level, the weight of $\frac{\{Z\}_L^l(i,j)}{\eta}$ is decreased from the coarsest level to the finest level. Clearly, the noise is prevented from being amplified in the sky region by using  the proposed multi-scale single image dehazing algorithm.

 {\it Case 2}: $\{t^*\}_G^l(i,j)$ is larger than $\eta$, $\phi(\{t^*\}_G^l(i,j))$ is almost 0.
  It can be derived from the equation (\ref{air1}) that
\begin{align}
\{I\}_L^l(i,j)\approx \frac{\{Z\}_L^l(i,j)}{\{t^*\}_G^l(i,j)}.
\end{align}
Subsequently, objects nearby the camera are restored well. It is shown in these two cases that single image dehazing is indeed a spatially varying single image detail enhancement \cite{Li2014}.

\section{Experimental Results}
\label{experiment}
Experimental results are provided in this section to demonstrate the proposed multi-scale dehazing algorithm. Readers are invited to view to electronic version of figures and zoom in them so as to better check differences among all images.

\subsection{Comparison of DCP and DDAP with Haze Line Averaging}

The DCP and the proposed DDAP with the haze line averaging are first compared through comparing the proposed dehazing algorithm  with the algorithm in \cite{Li2015} which is on top of the DCP. To reduce the possible morphological artifacts generated by the DCP, the radius of the WGIF \cite{Li2014} is selected as 60. The algorithm in \cite{Li2015} is also extended to two scales. The WGIF is a local patch-based edge-preserving smoothing filter and it could be over-smooth in the areas of fine structure. As a result, details in such areas cannot be preserved well. By zooming in Fig. \ref{Fig3},  the hair can be found to be over-smooth by the WGIF if the radius is selected as 60.  The problem can be alleviated by using the two scales and the proposed DDAP with the haze line averaging. In addition, there are visibly morphological artifacts in the Fig. \ref{Fig3}(a) and Fig. \ref{Fig3}(b) while they are significantly reduced in Fig. \ref{Fig3}(c) and Fig. \ref{Fig3}(d).

\begin{figure}[!htb]
\centering{
\subfigure[ $\pi/120\times \pi/120$]{\includegraphics[width=1.6in]{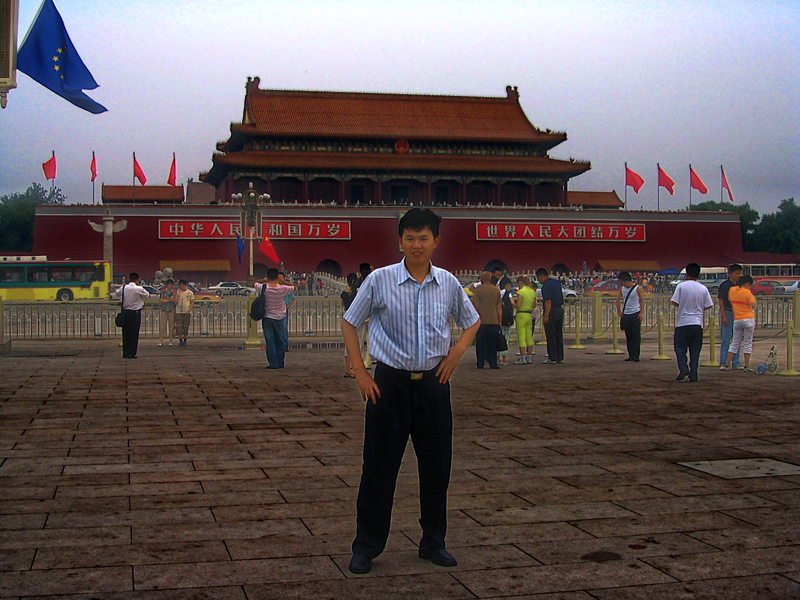}}
\subfigure[ $\pi/360\times \pi/360$]{\includegraphics[width=1.6in]{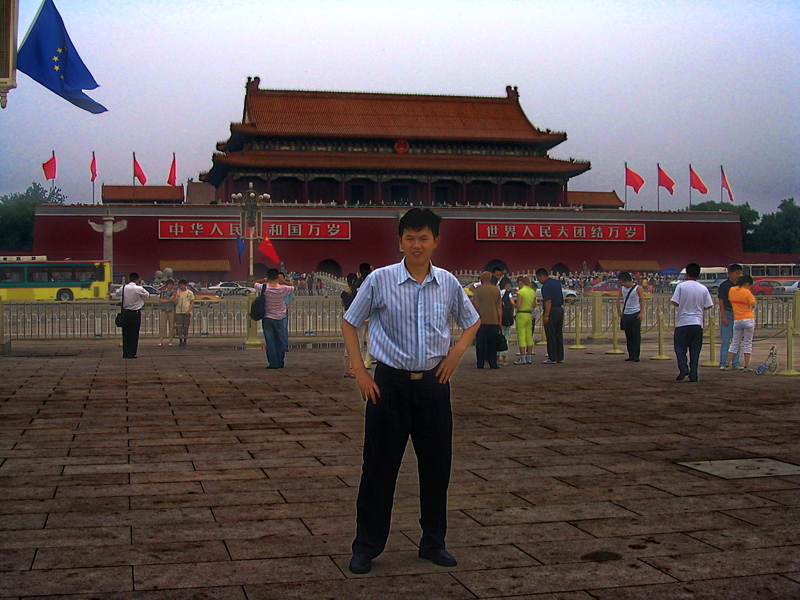}}
}
\caption{Comparison of  bin sizes for the haze line averaging.}
\label{Fig36}
\end{figure}

\begin{figure}[!htb]
\centering{
\subfigure[ $\nu$=50]{\includegraphics[width=1.6in]{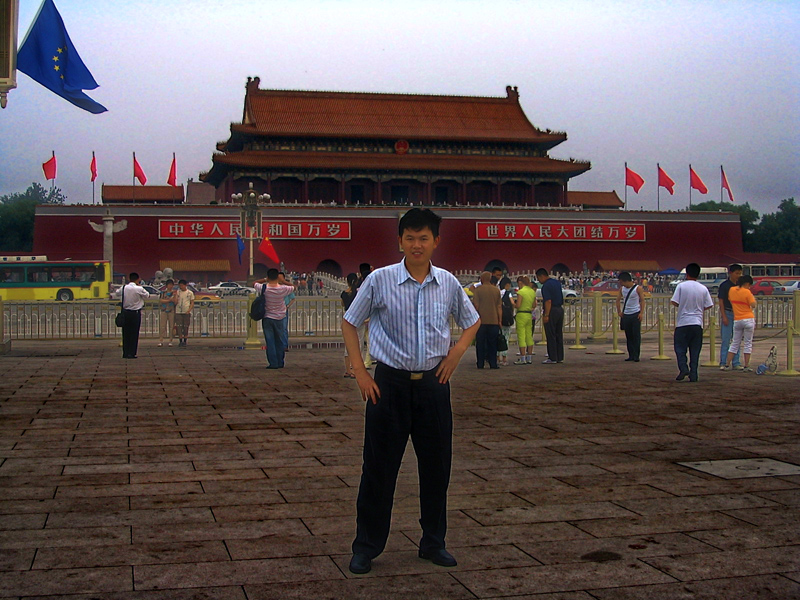}}
\subfigure[ $\nu$=200]{\includegraphics[width=1.6in]{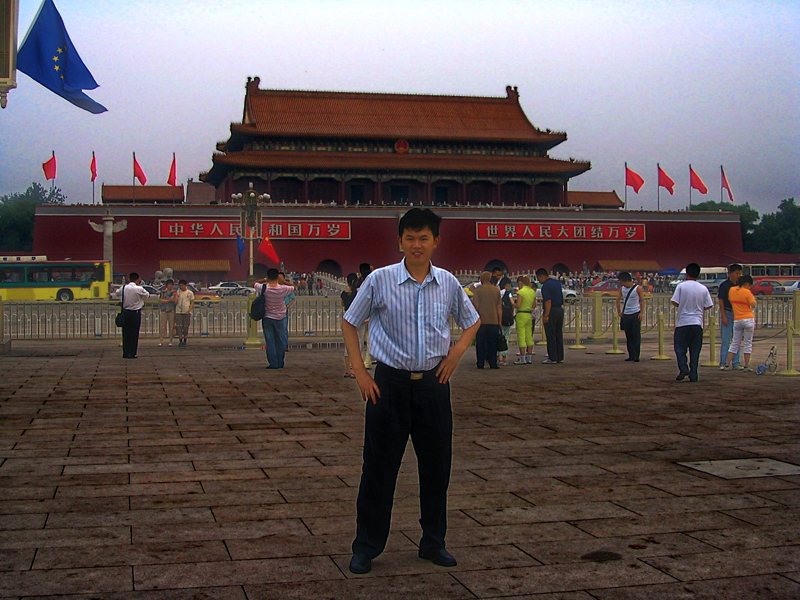}}
}
\caption{Comparison of different $\nu$'s for the haze line averaging.}
\label{Fig66}
\end{figure}

Both the bin size and the $\nu$ are important for the proposed haze line averaging. Besides the recommended bin size, two more sizes of $\pi/120\times \pi/120$ and $\pi/360\times \pi/360$ are tested. As shown in Fig. \ref{Fig36}, the color is slightly distorted if the bin size is selected as $\pi/120\times \pi/120$ while the color distortion is reduced by choosing the bin size as $\pi/360\times \pi/360$. It is thus recommended that the bin size of $\theta$ or $\psi$ is smaller than $\pi/360$. Two more choices of $\nu$ as 50 and 200 are also tested. As illustrated in Fig. \ref{Fig66}, there are slightly visible morphological artifacts by choosing the $\nu$ as 50 while the morphological artifacts are removed by selecting the $\nu$ as 200.

\subsection{Comparison of Different Dehazing Algorithms}

The proposed multi-scale algorithm is  compared with eight state-of-the-art dehazing algorithms in  DehazeNet \cite{cai}, EPDN \cite{Qu_2019_CVPR}, FFA-Net \cite{1qin2020}, MSBDN \cite{1dong2020},  DCP \cite{1he2013}, HLP \cite{CVPR16},  AOD \cite{ICCV17},  and \cite{Li2021}. Among them, the algorithms in \cite{1dong2020} and \cite{Li2021} are also multi-scale ones.  It should be pointed out that the algorithms in \cite{CVPR16}, \cite{1he2013},  and \cite{Li2021} are model-based algorithms while the others are data-driven algorithms.

\begin{figure*}[!htb]
\centering{
{\includegraphics[width=0.7in]{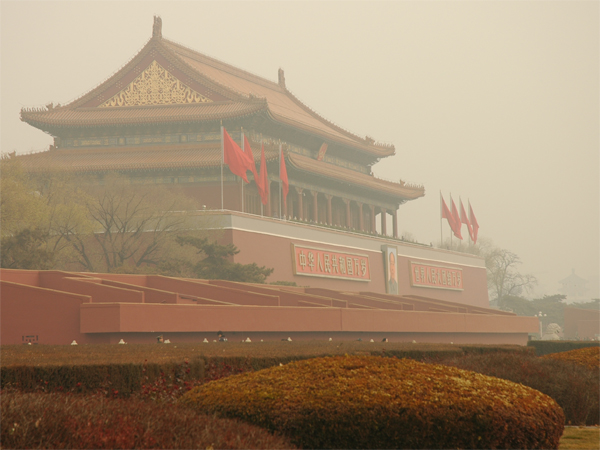}}
{\includegraphics[width=0.7in]{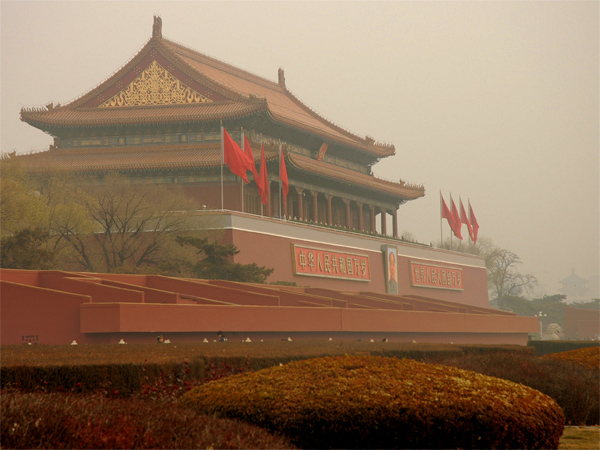}}
{\includegraphics[width=0.7in]{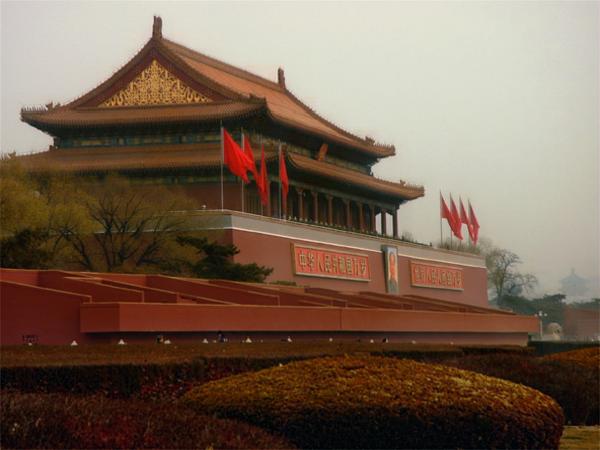}}
{\includegraphics[width=0.7in]{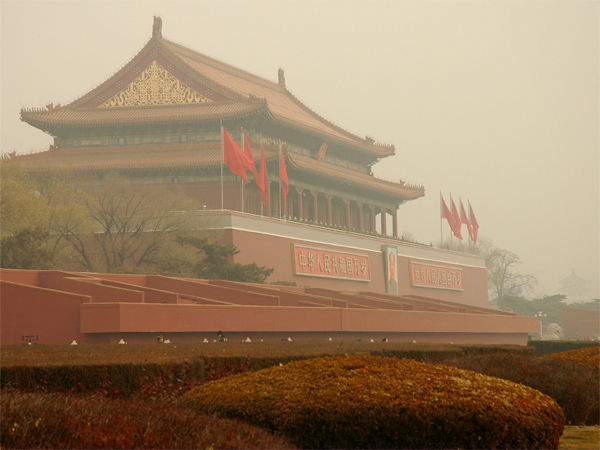}}
{\includegraphics[width=0.7in]{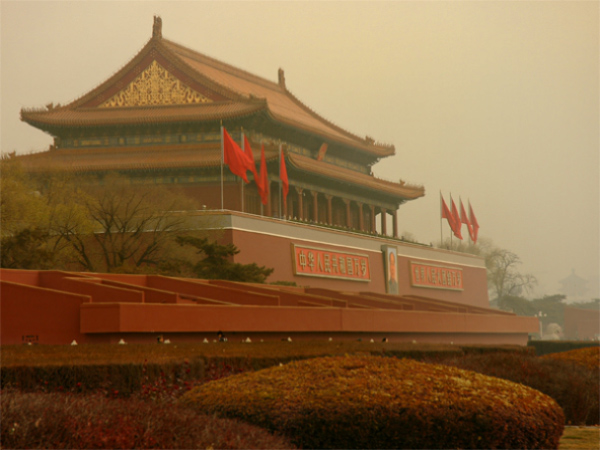}}
{\includegraphics[width=0.7in]{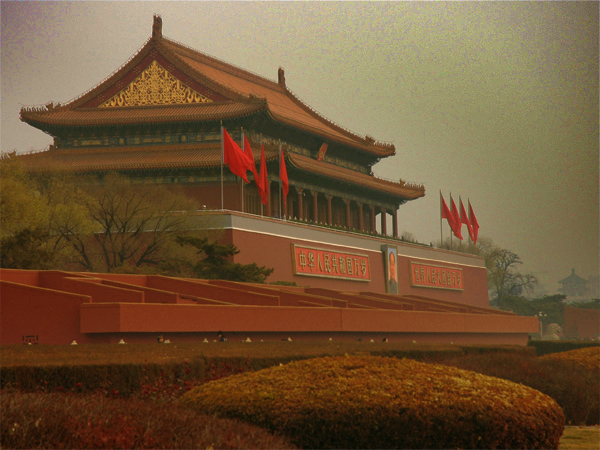}}
{\includegraphics[width=0.7in]{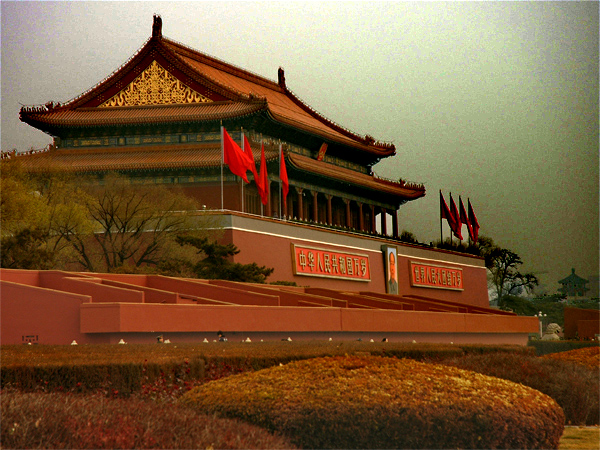}}
{\includegraphics[width=0.7in]{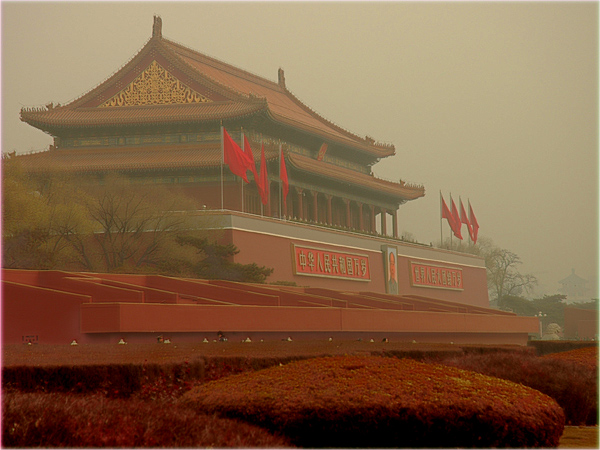}}
{\includegraphics[width=0.7in]{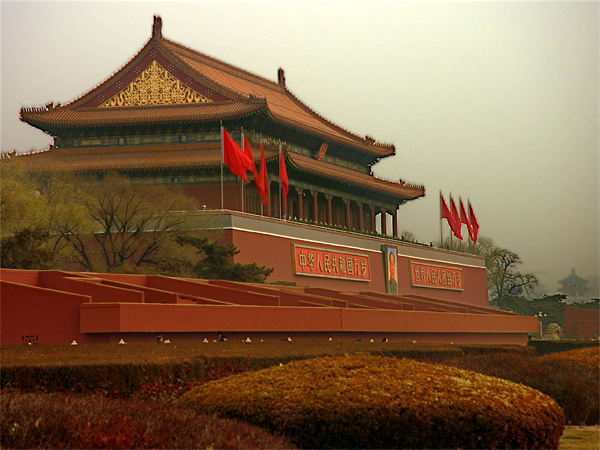}}
}

\centering{
{\includegraphics[width=0.7in]{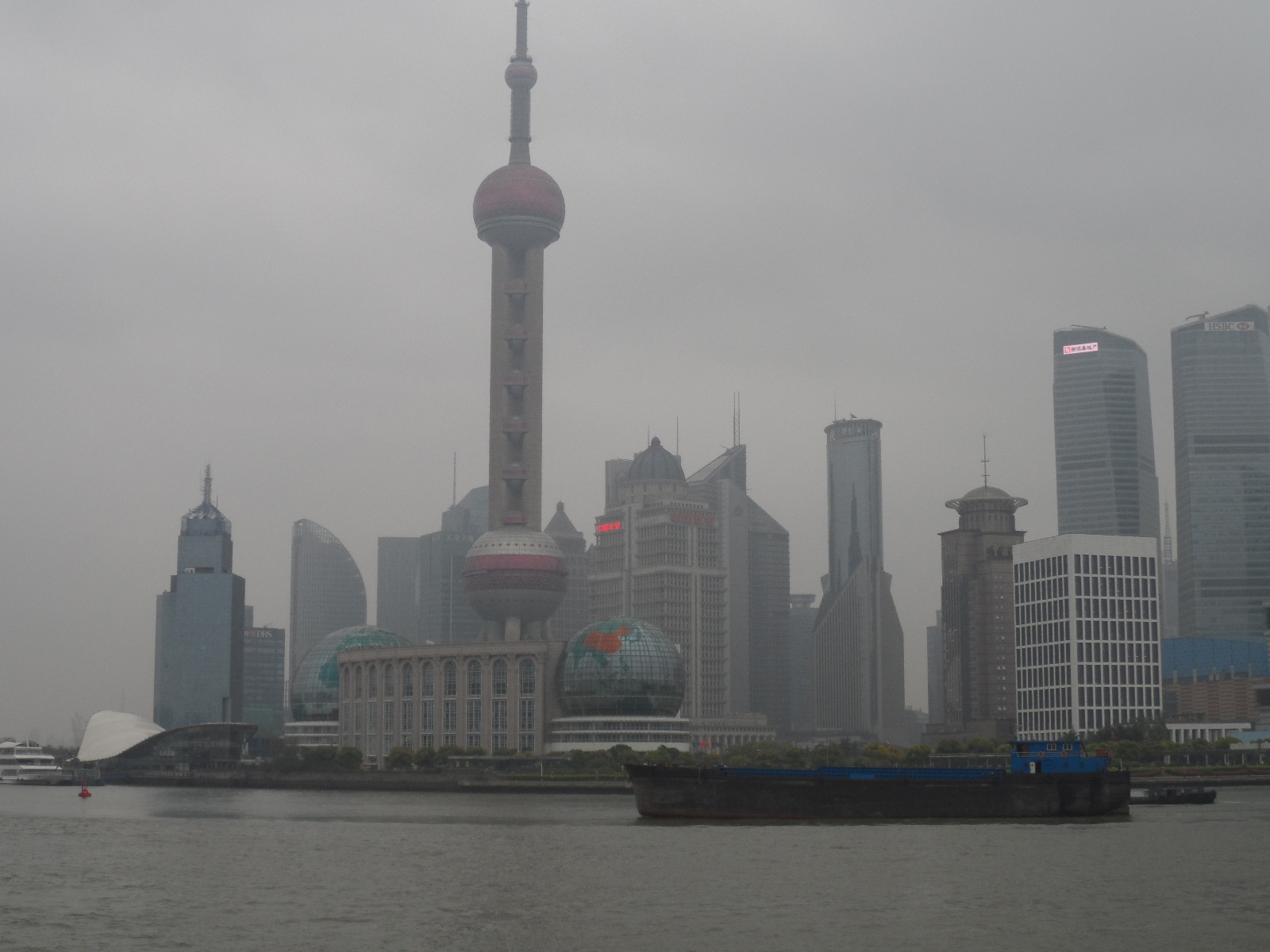}}
{\includegraphics[width=0.7in]{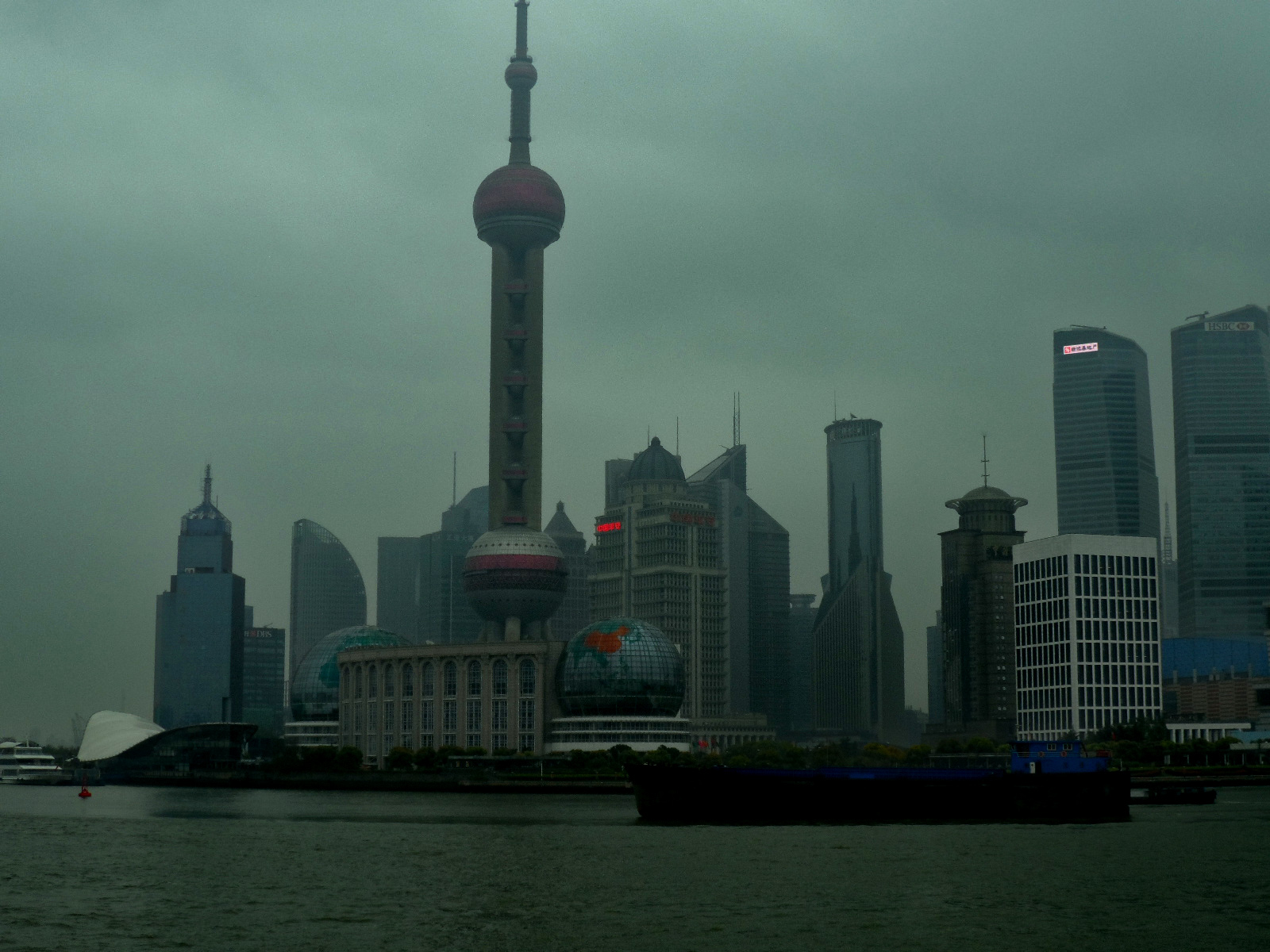}}
{\includegraphics[width=0.7in]{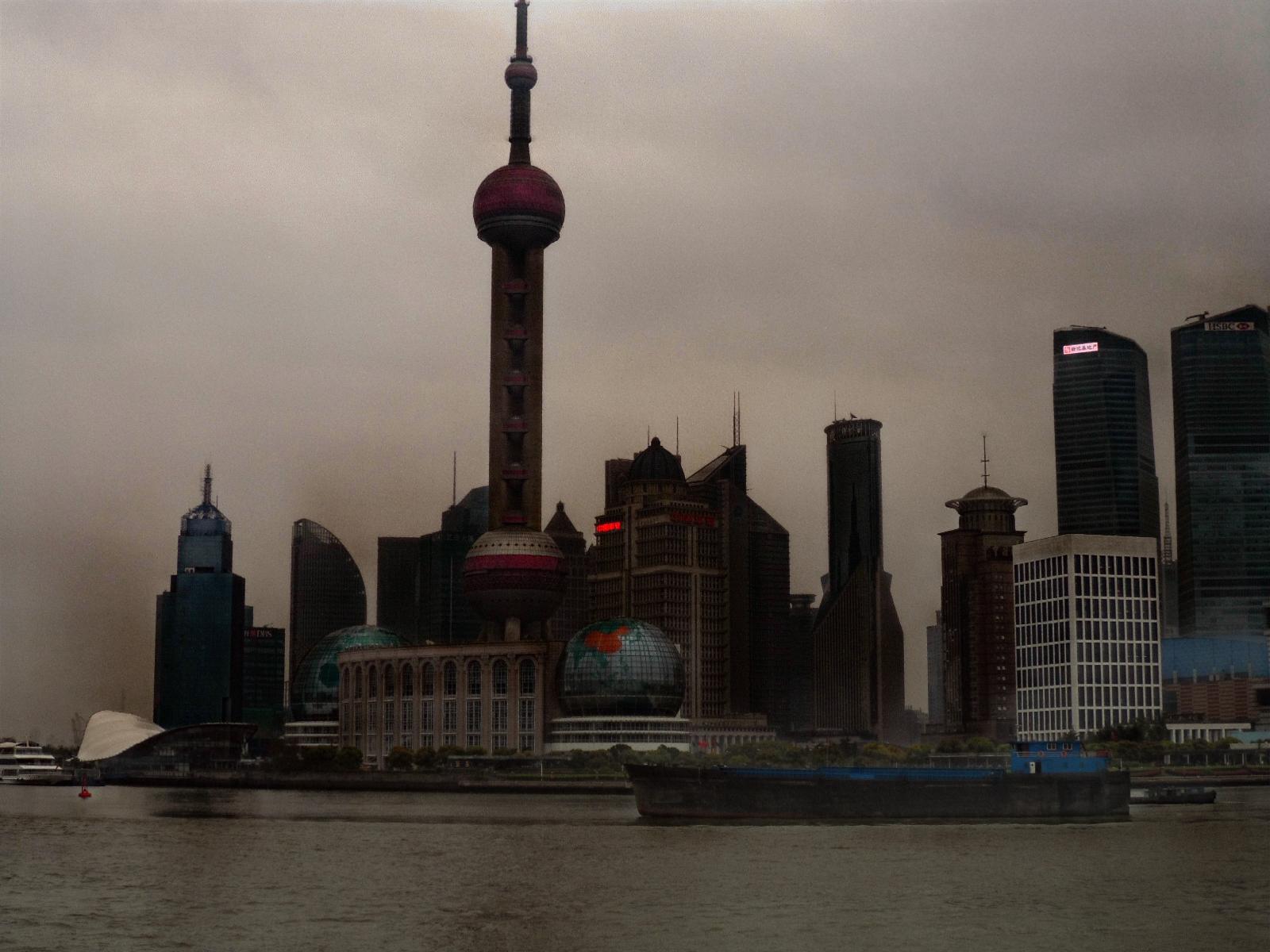}}
{\includegraphics[width=0.7in]{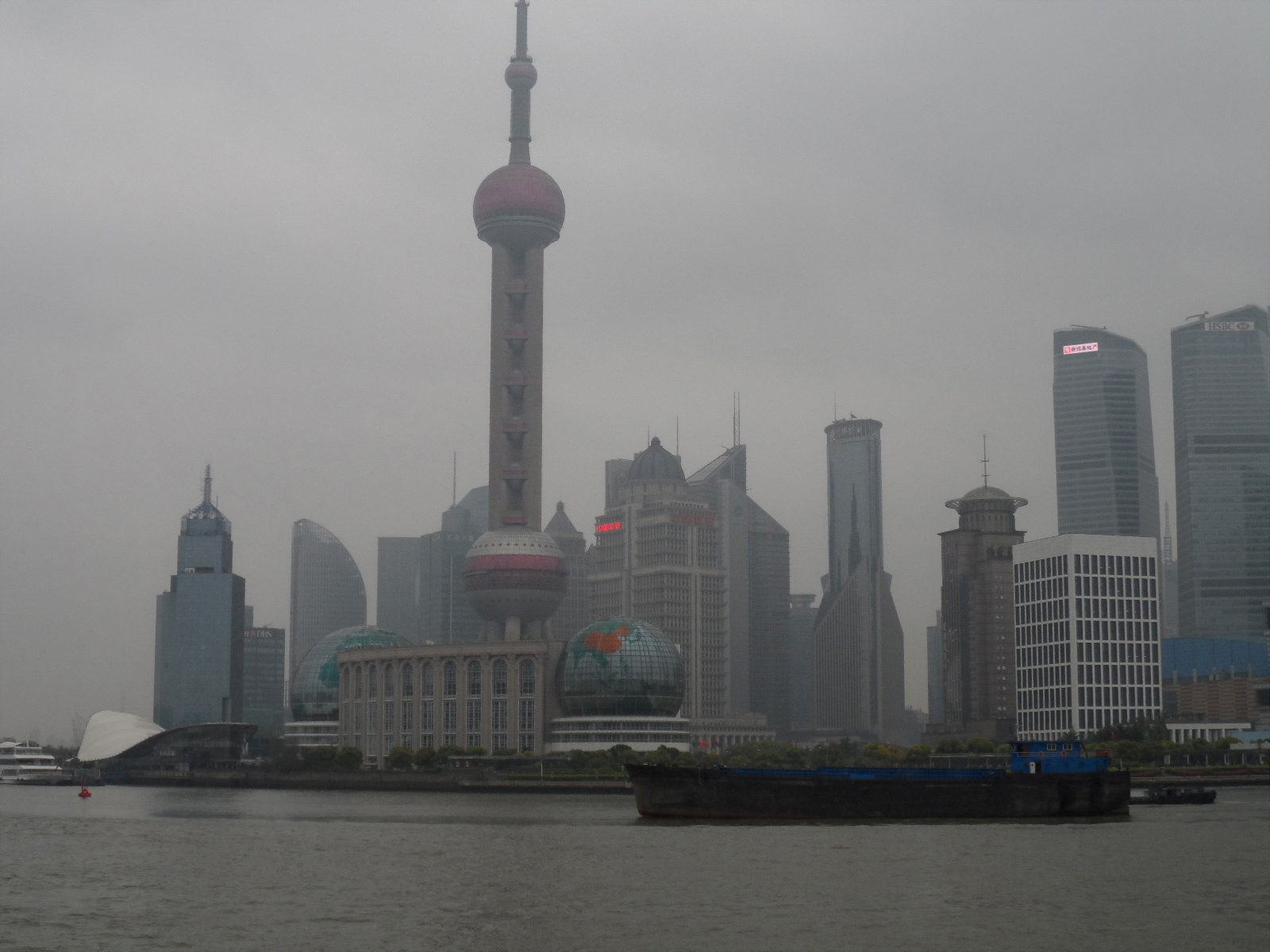}}
{\includegraphics[width=0.7in]{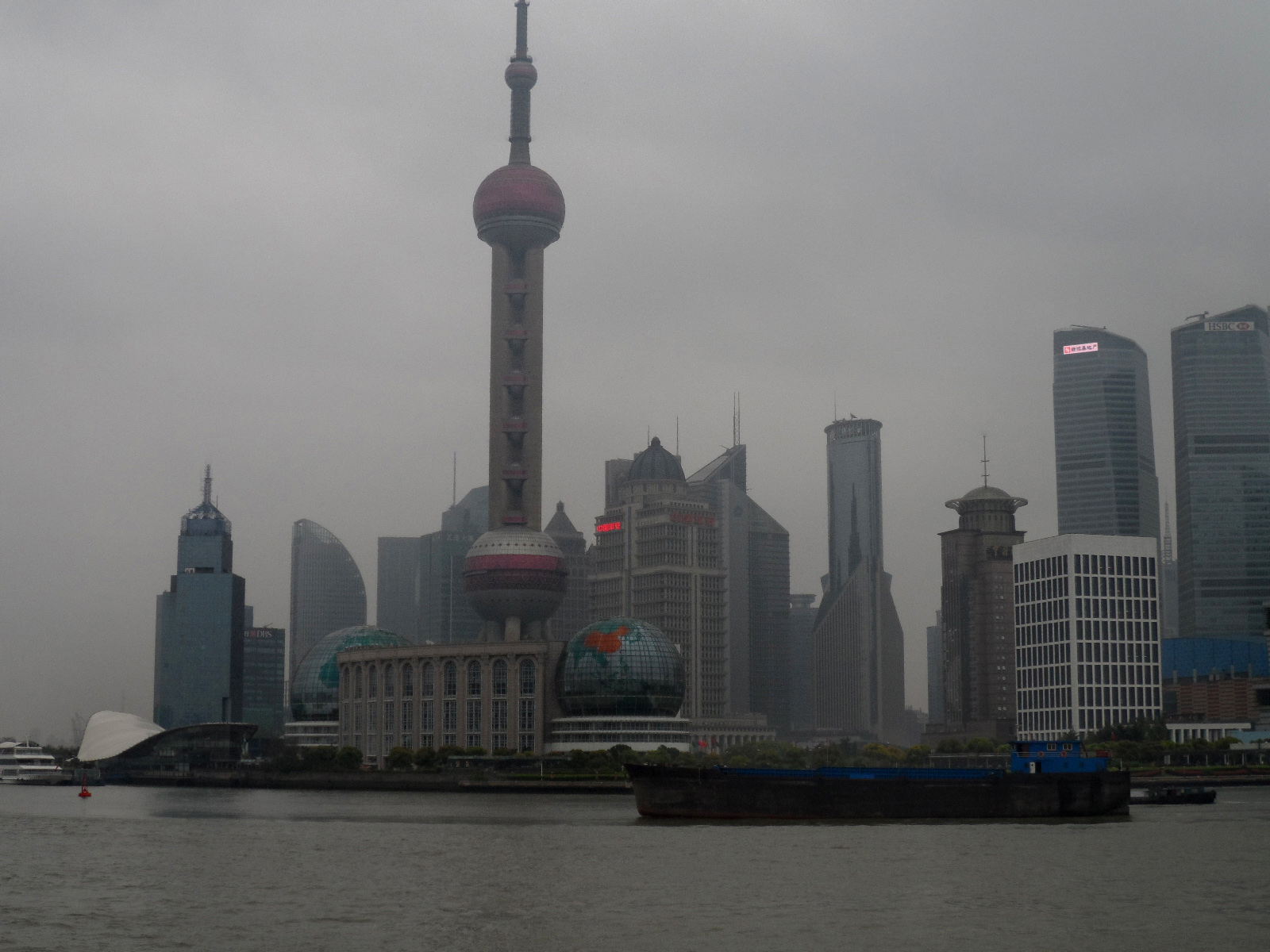}}
{\includegraphics[width=0.7in]{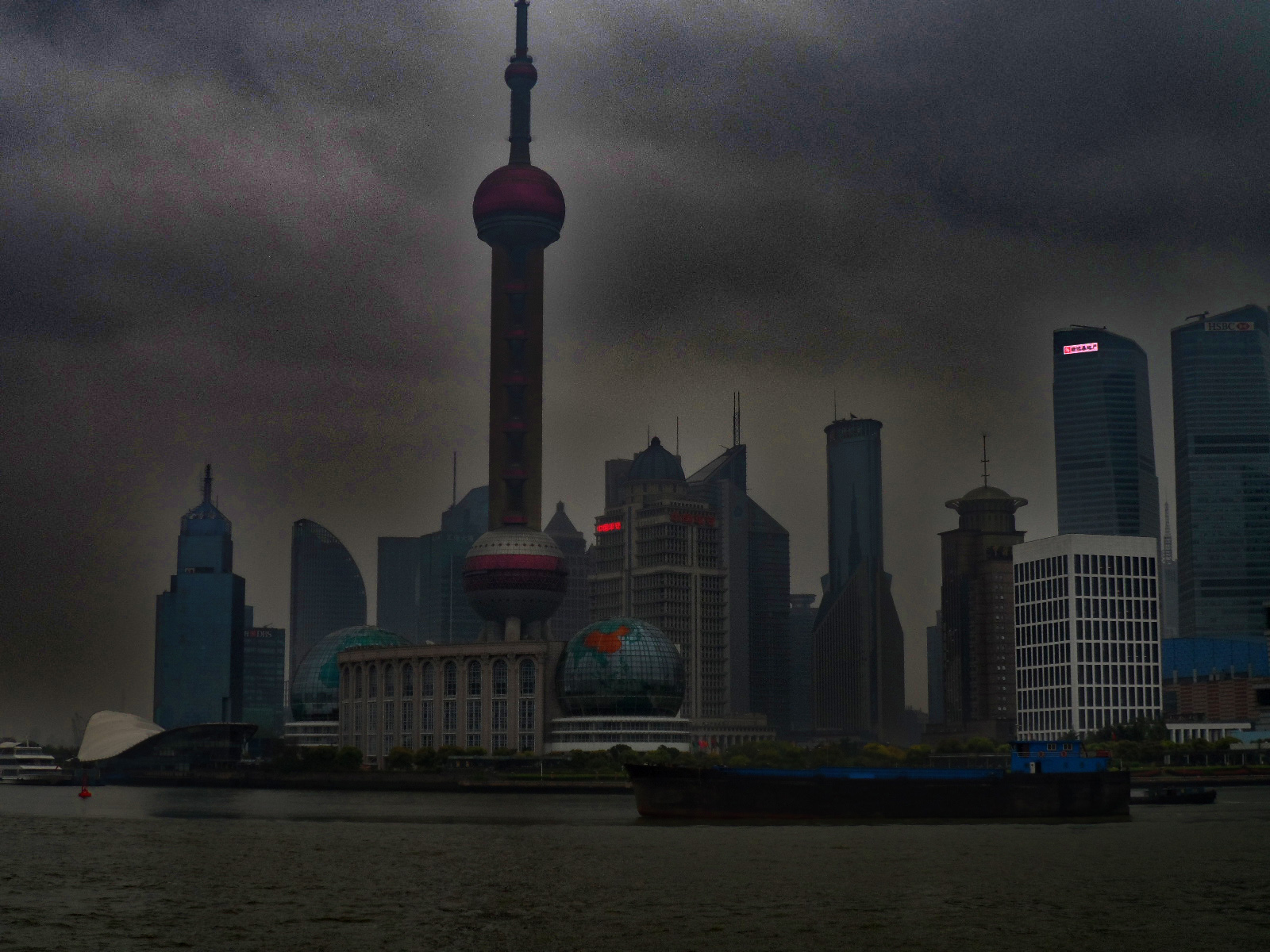}}
{\includegraphics[width=0.7in]{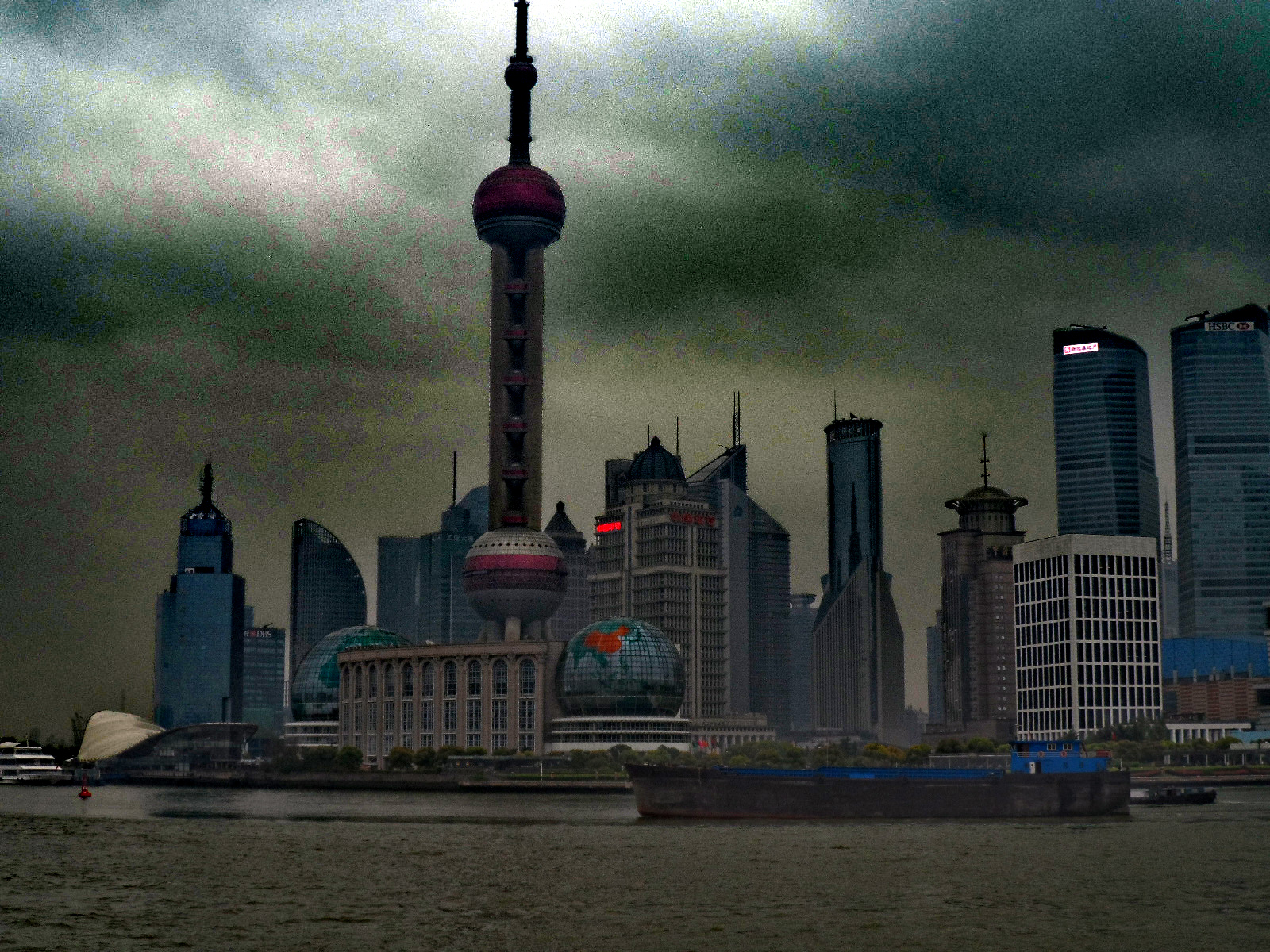}}
{\includegraphics[width=0.7in]{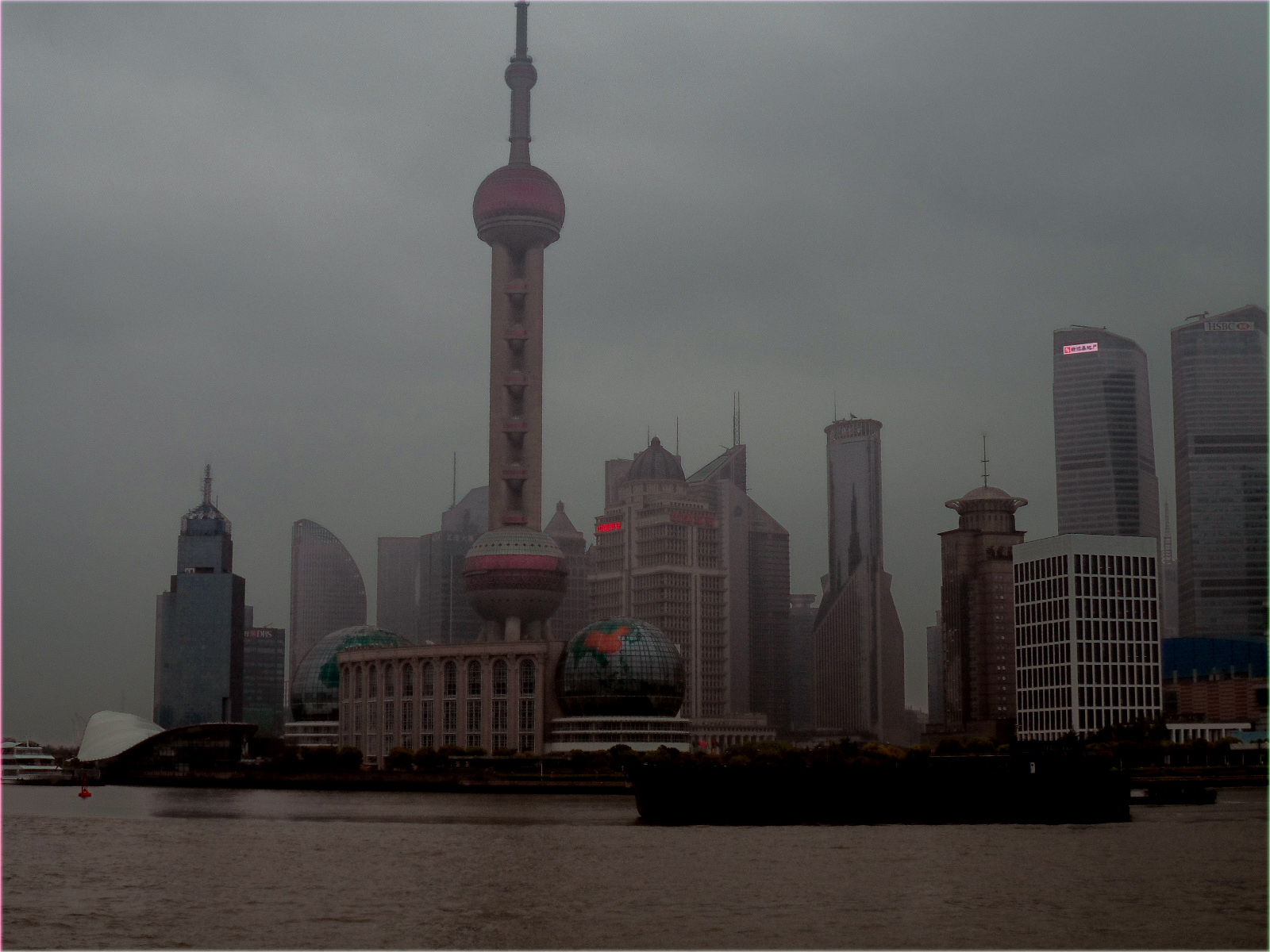}}
{\includegraphics[width=0.7in]{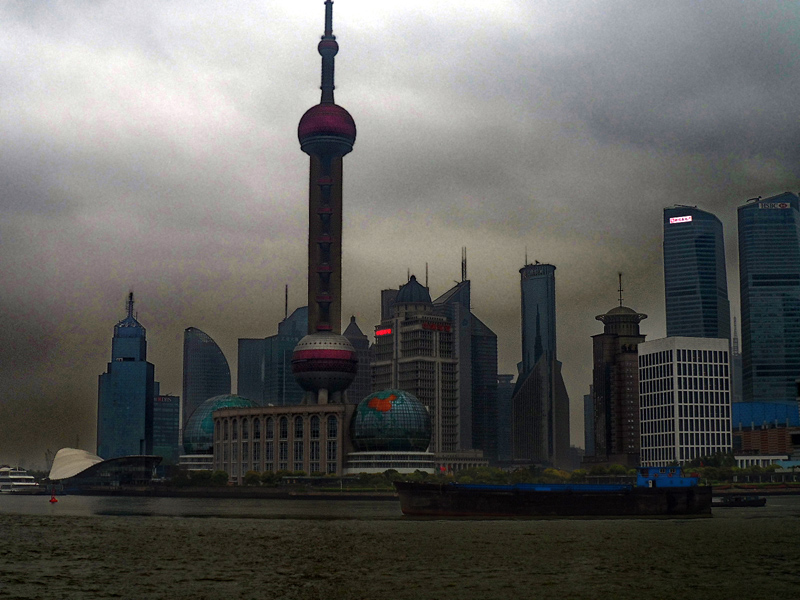}}	
}

\centering{
{\includegraphics[width=0.7in]{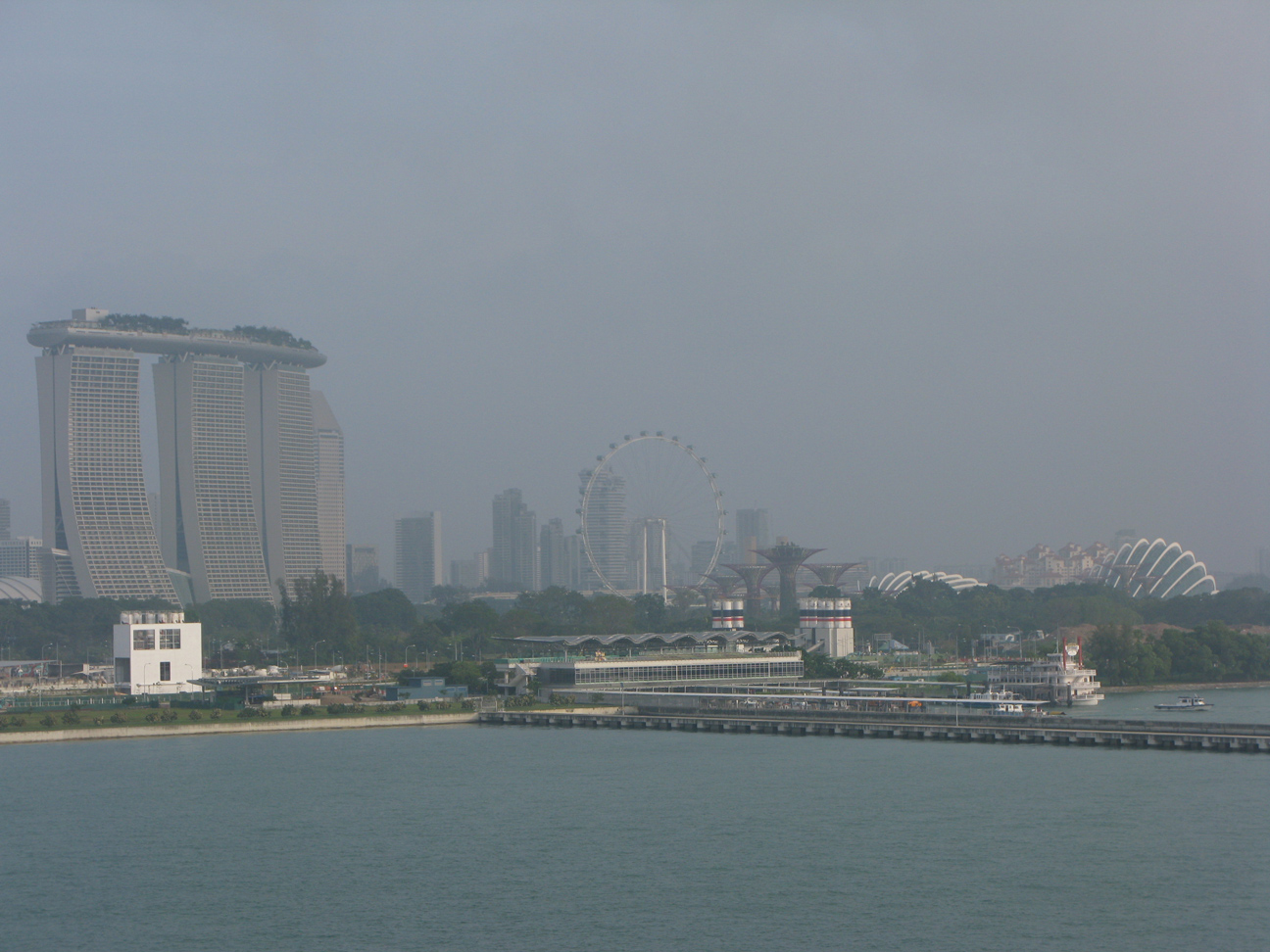}}
{\includegraphics[width=0.7in]{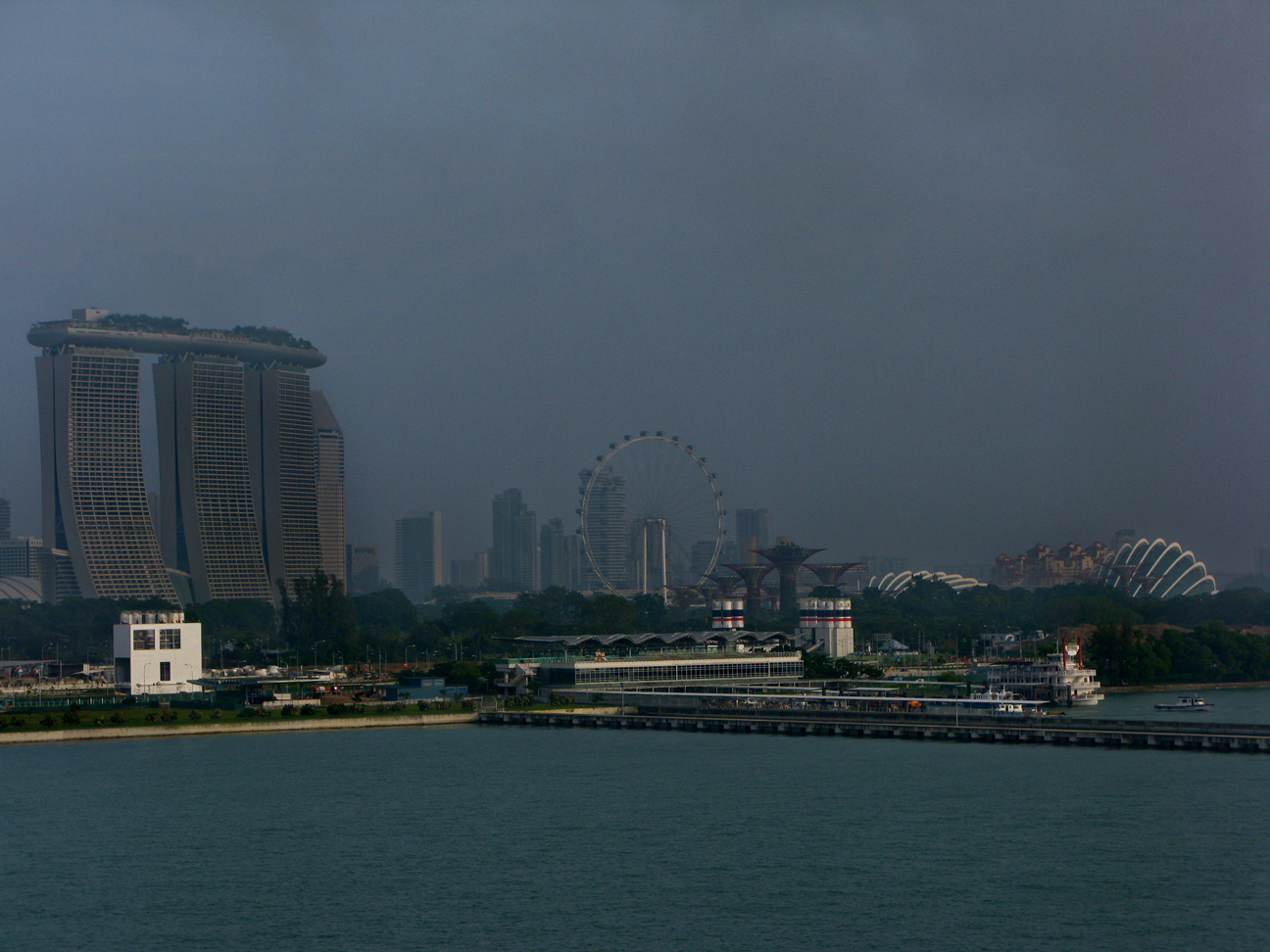}}
{\includegraphics[width=0.7in]{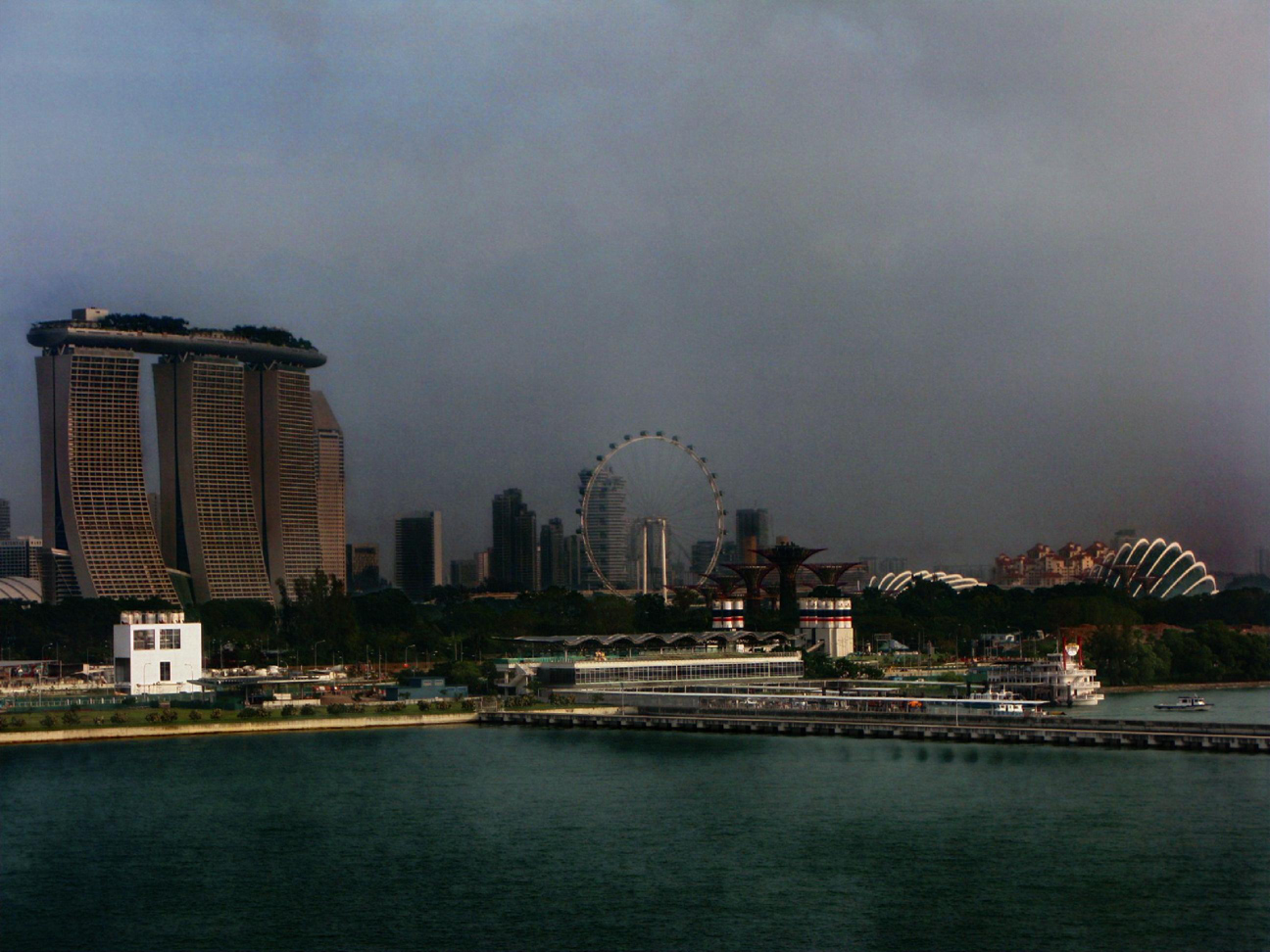}}
{\includegraphics[width=0.7in]{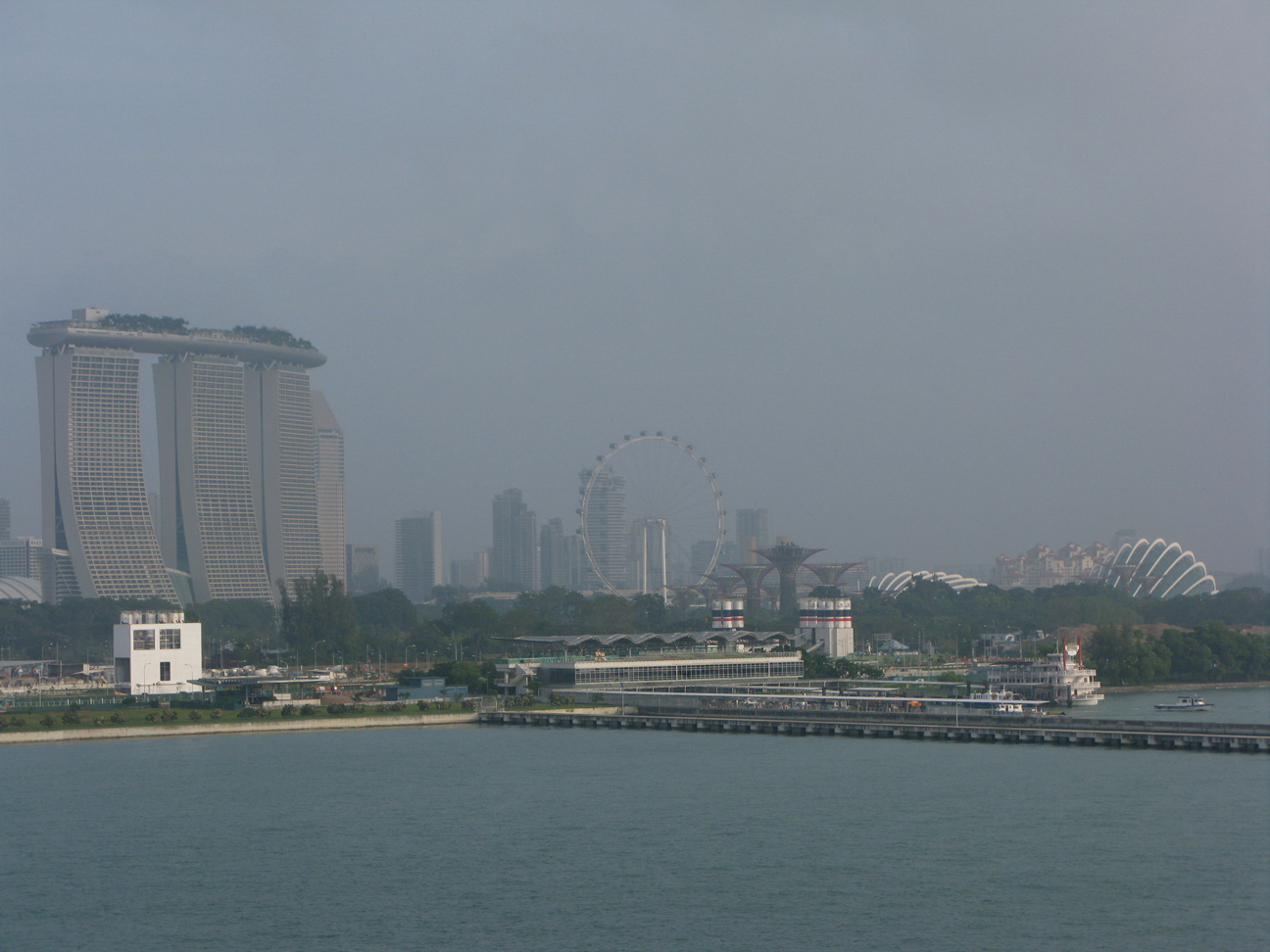}}
{\includegraphics[width=0.7in]{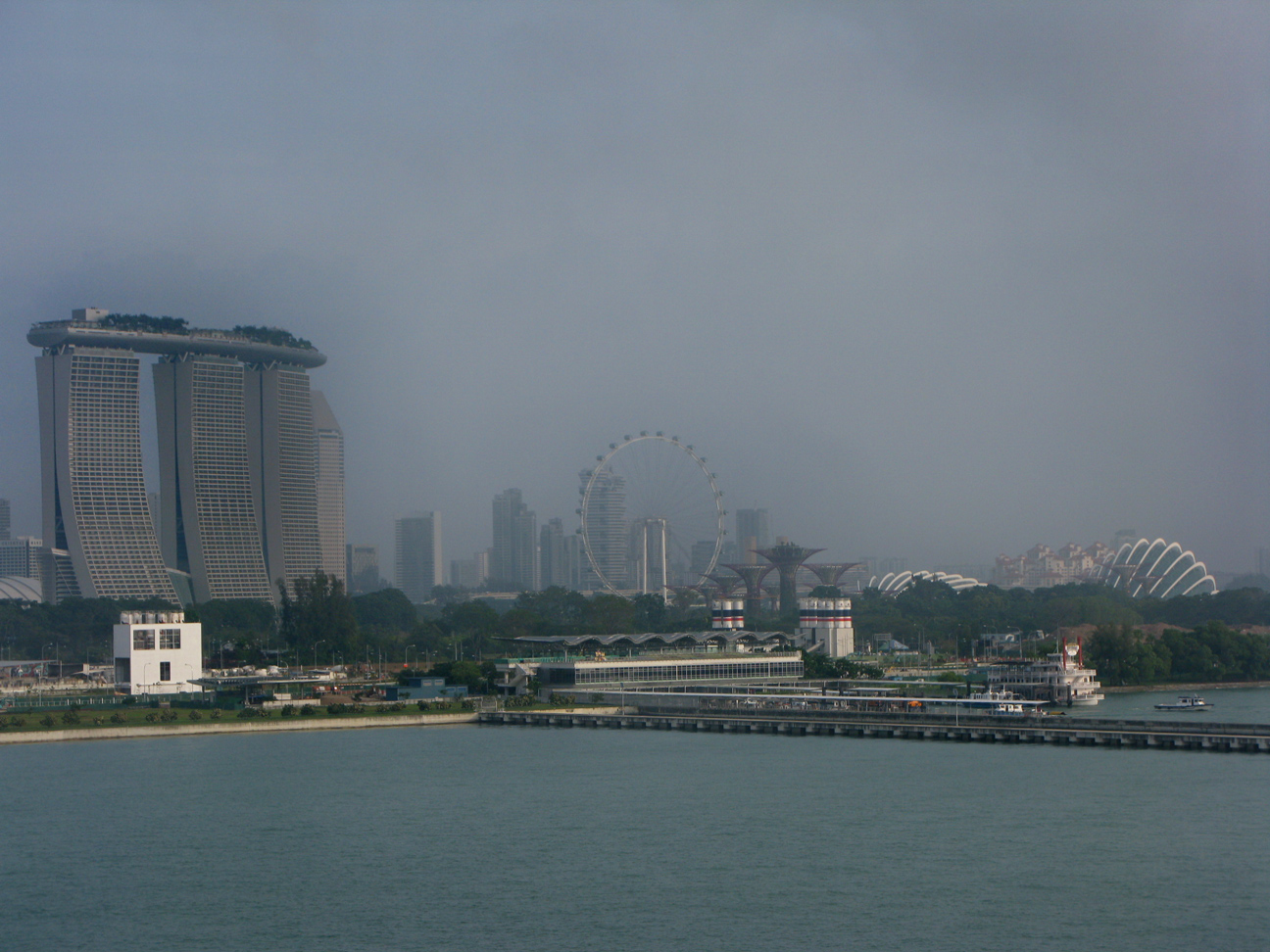}}
{\includegraphics[width=0.7in]{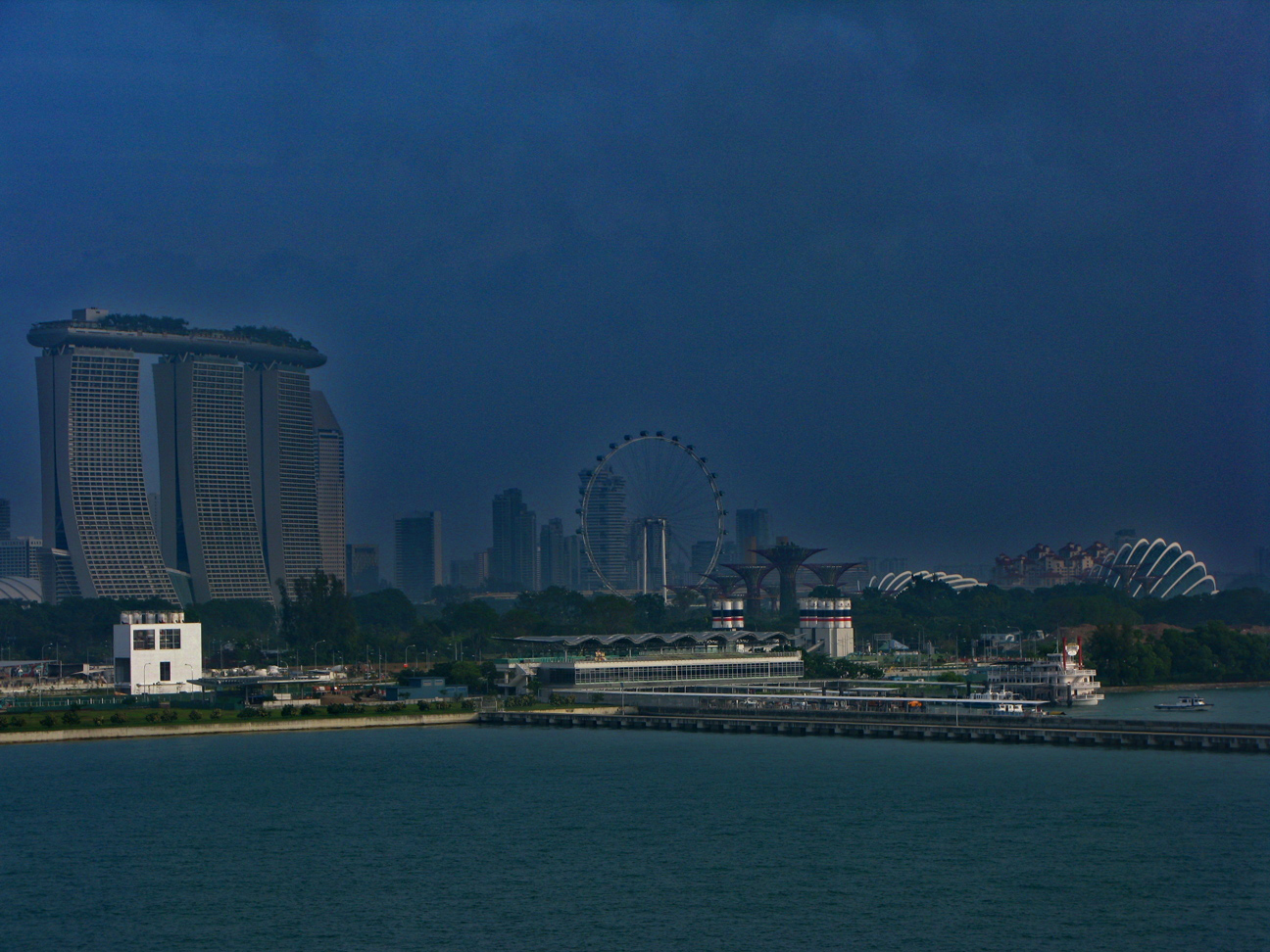}}
{\includegraphics[width=0.7in]{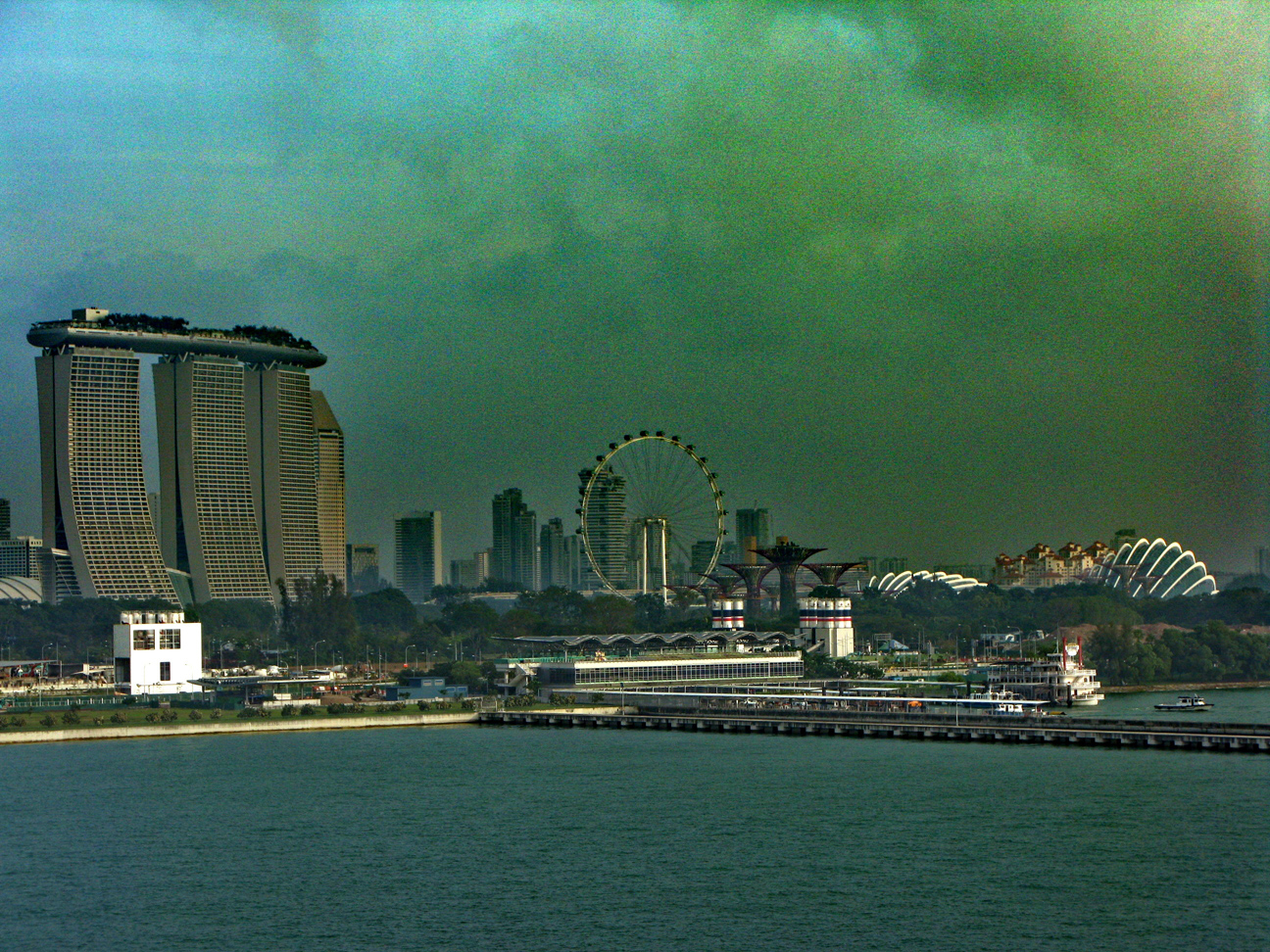}}
{\includegraphics[width=0.7in]{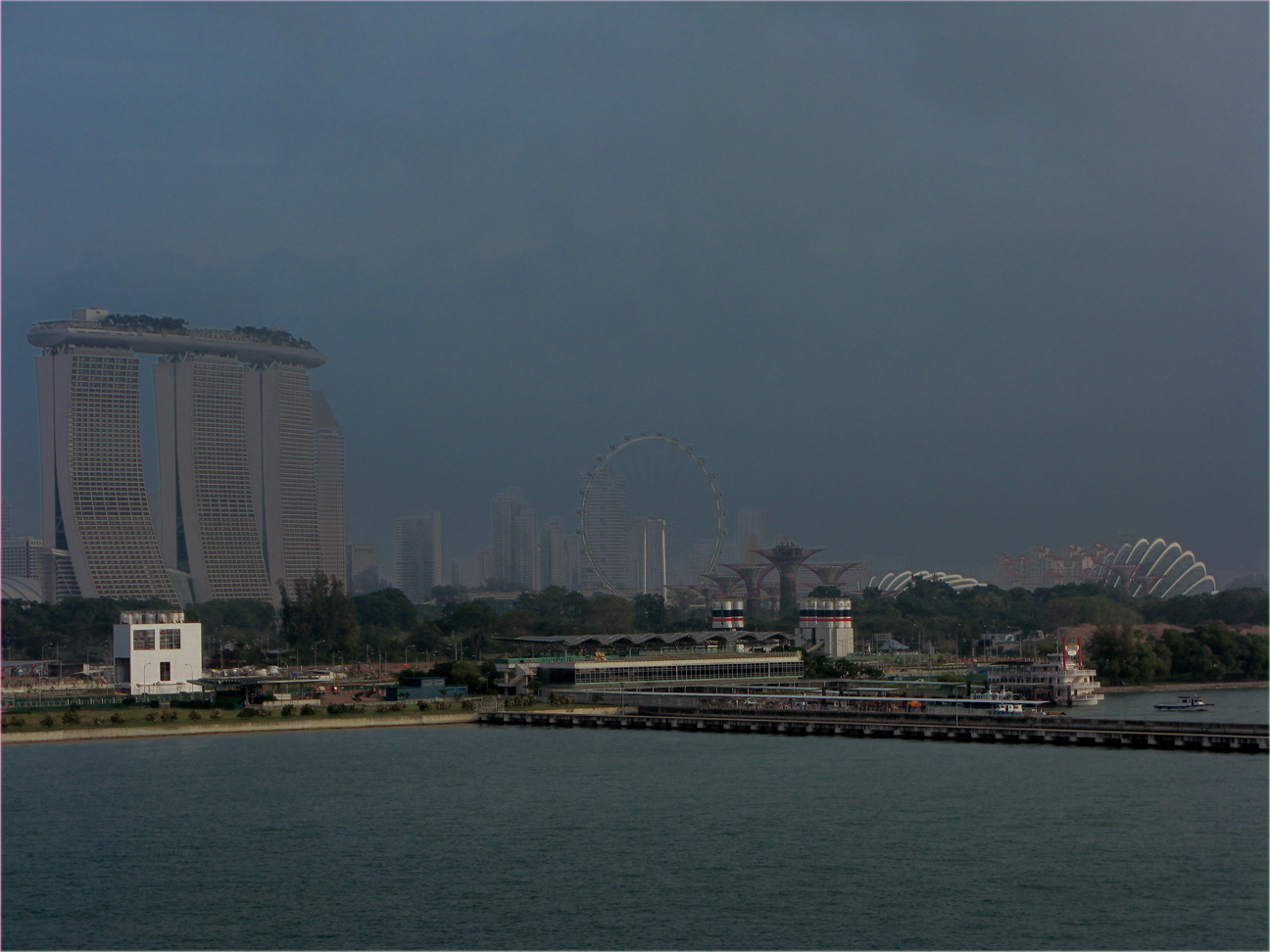}}
{\includegraphics[width=0.7in]{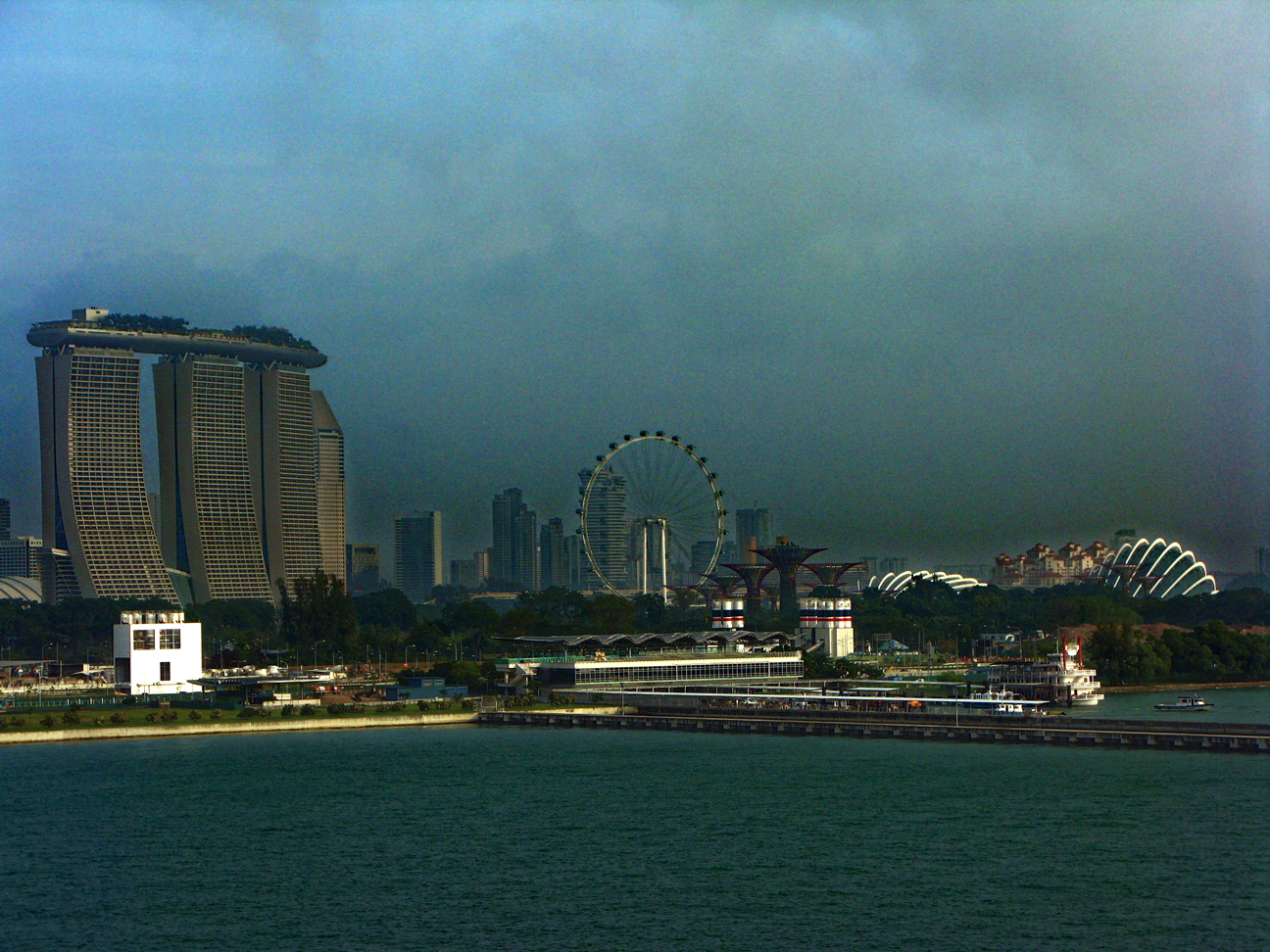}}	
}

\centering{
\subfigure[Input]{\includegraphics[width=0.7in]{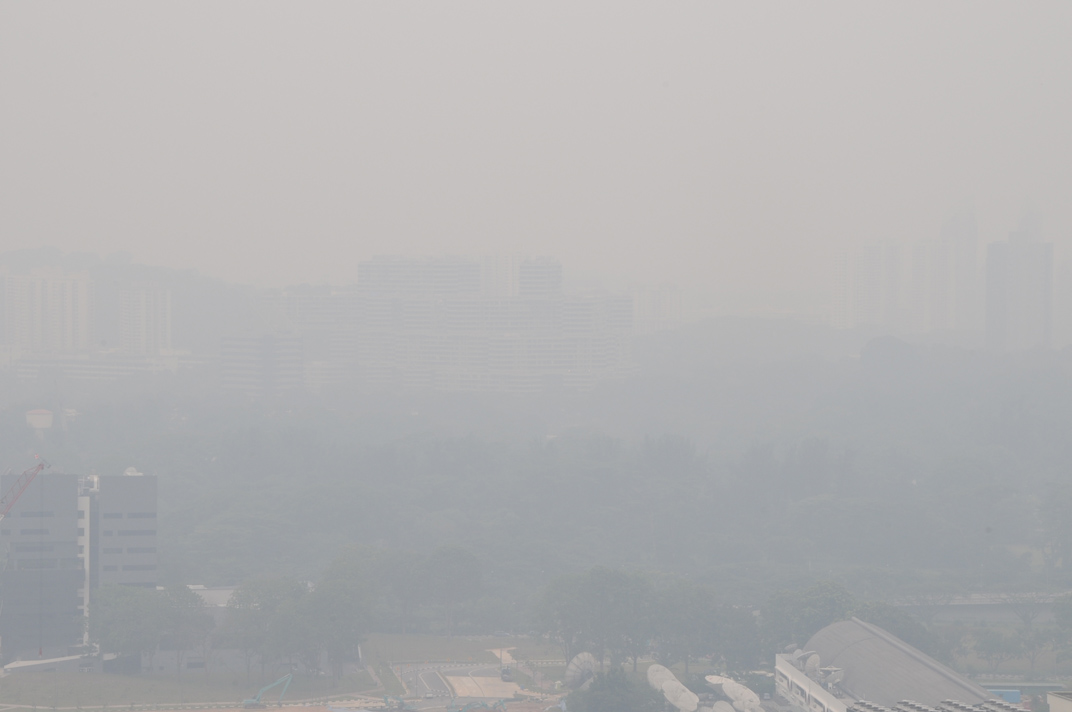}}
\subfigure[DehazeNet]{\includegraphics[width=0.7in]{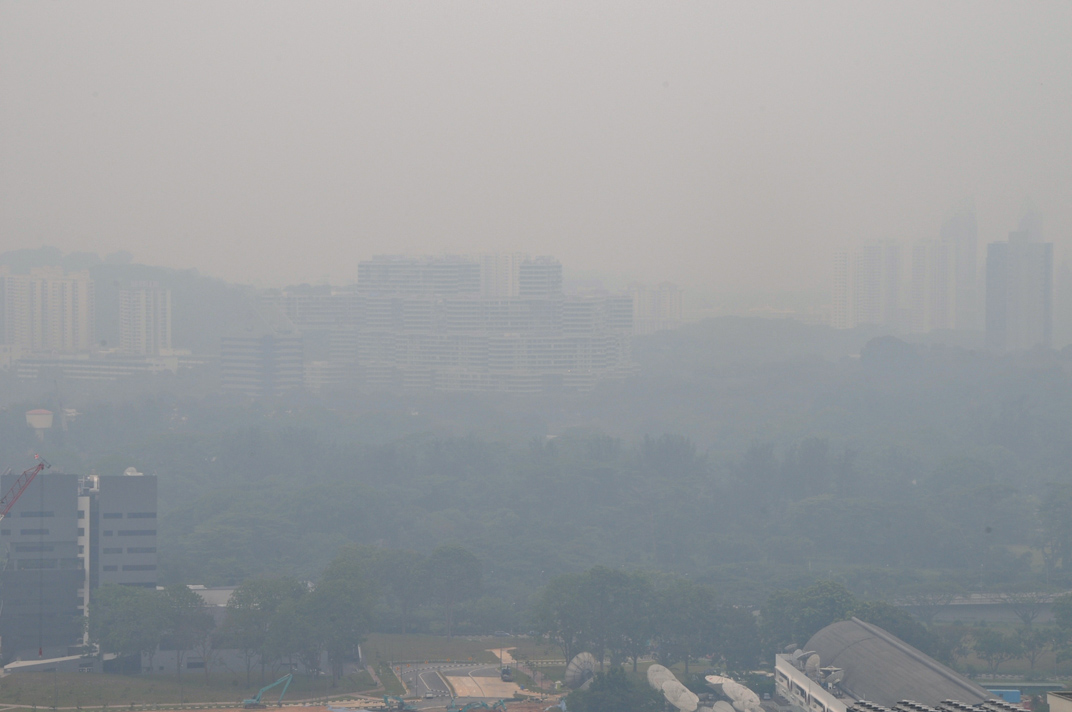}}
\subfigure[EPDN]{\includegraphics[width=0.7in]{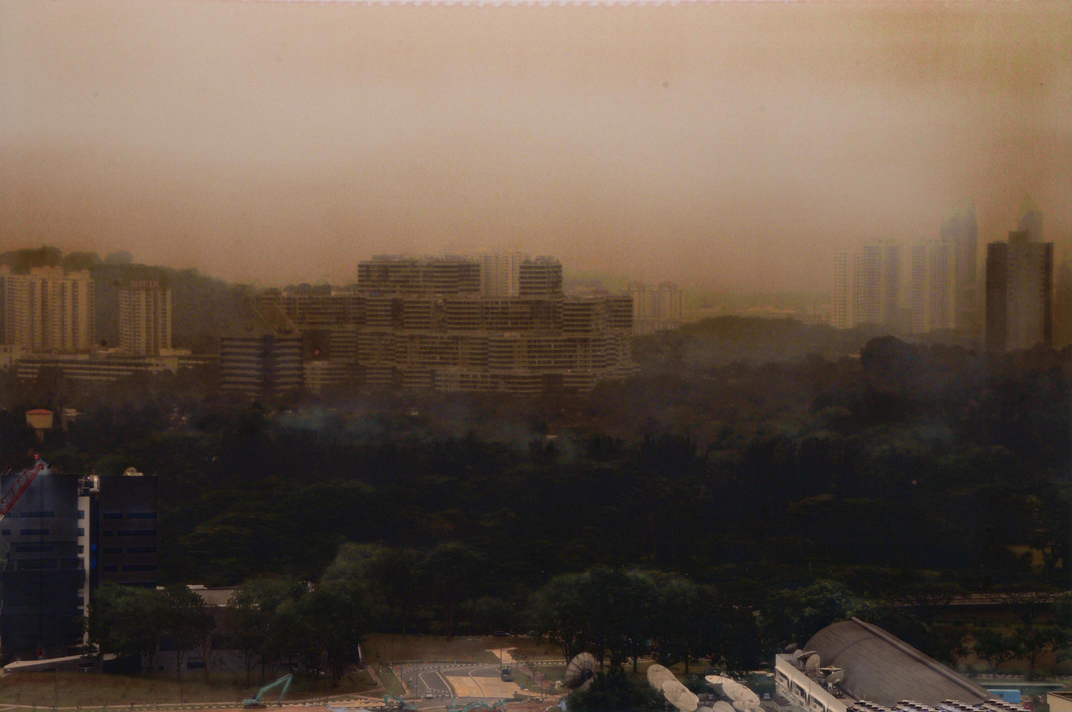}}
\subfigure[FFA-Net]{\includegraphics[width=0.7in]{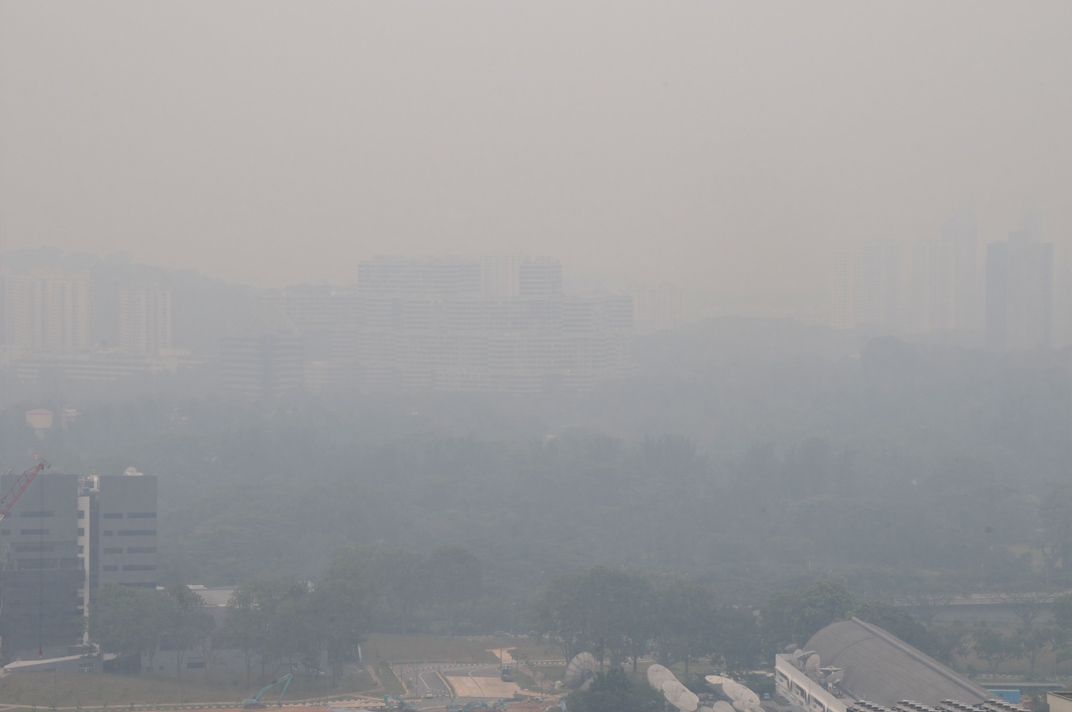}}
\subfigure[MSBDN]{\includegraphics[width=0.7in]{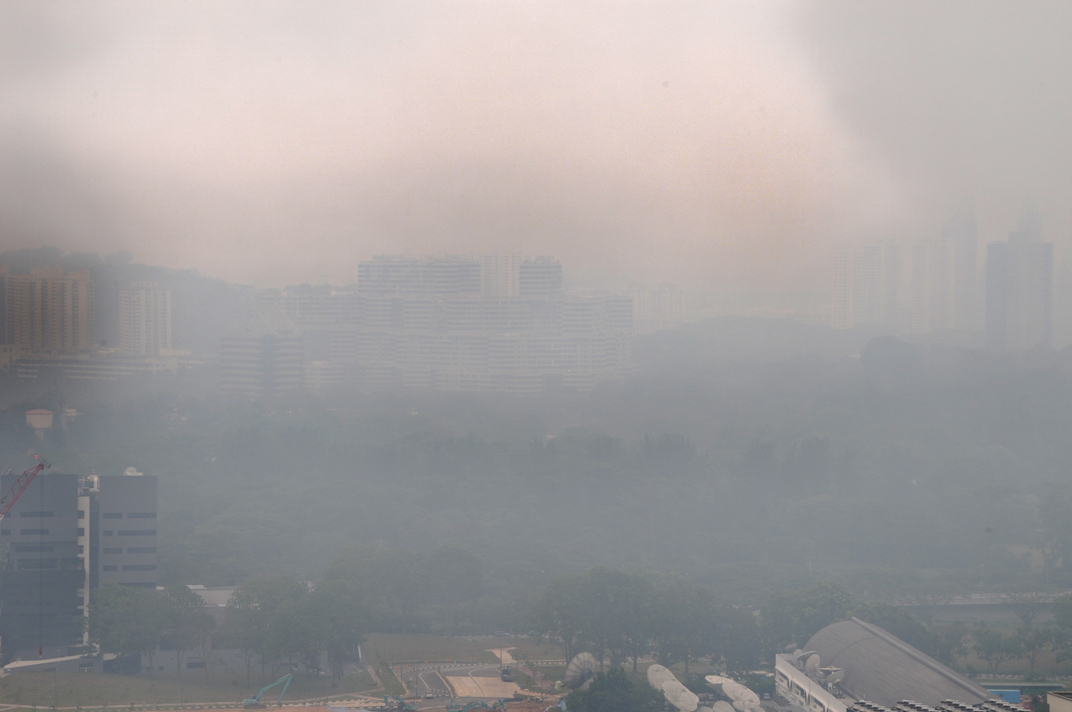}}
\subfigure[DCP]{\includegraphics[width=0.7in]{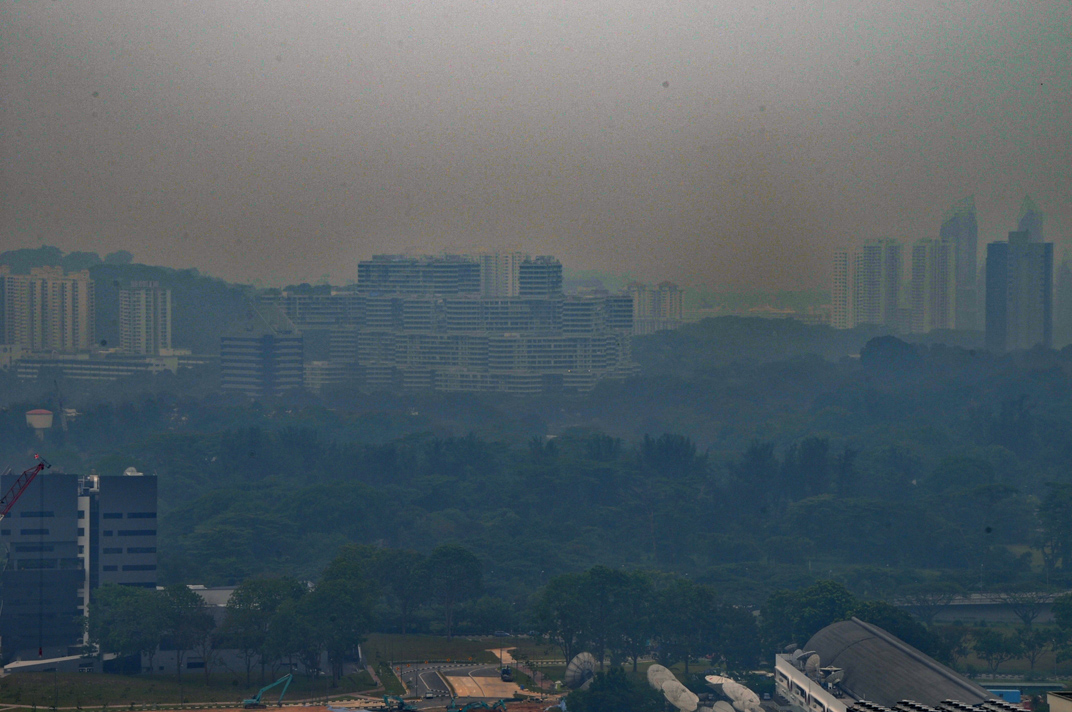}}
\subfigure[HLP]{\includegraphics[width=0.7in]{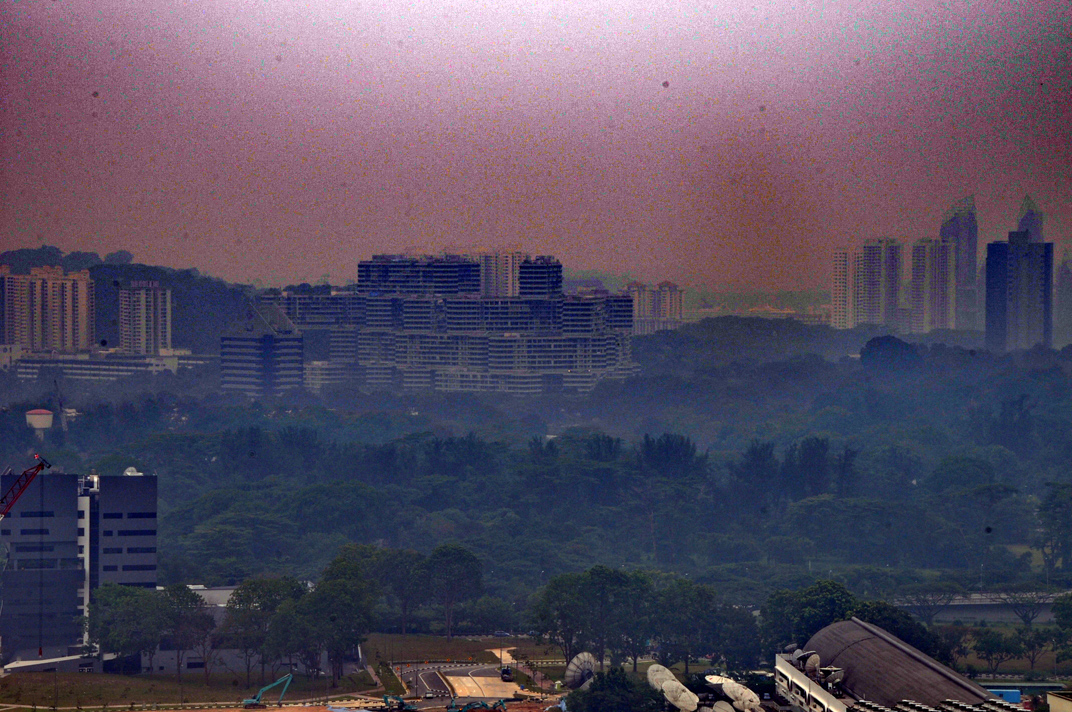}}
\subfigure[AOD]{\includegraphics[width=0.7in]{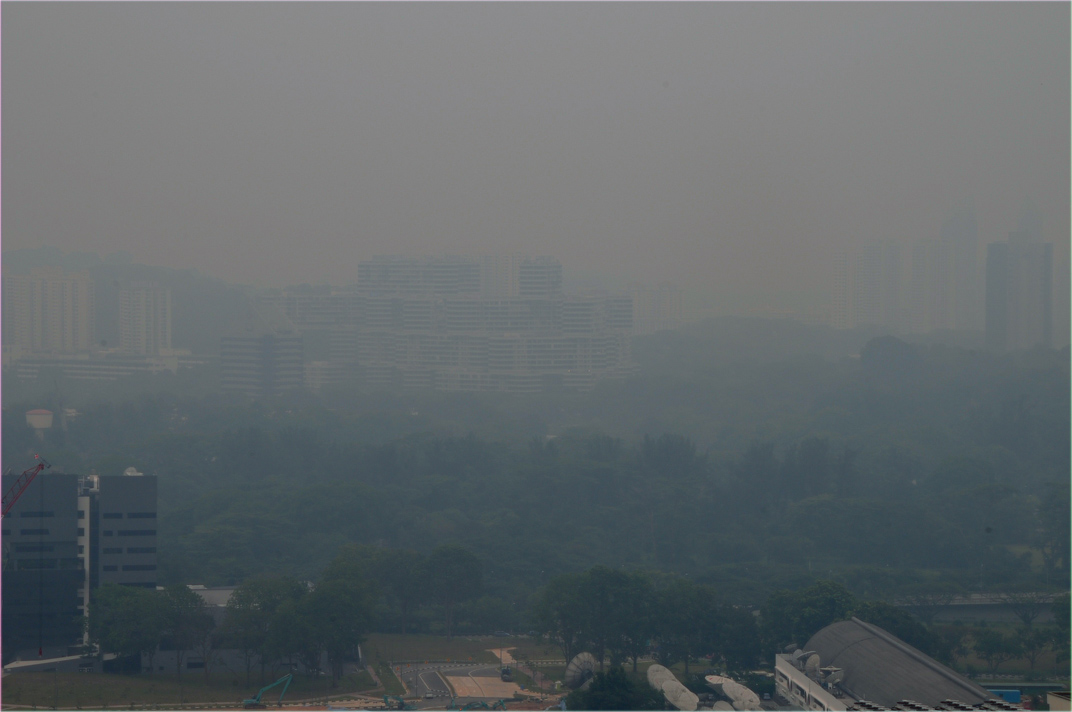}}
\subfigure[Ours]{\includegraphics[width=0.7in]{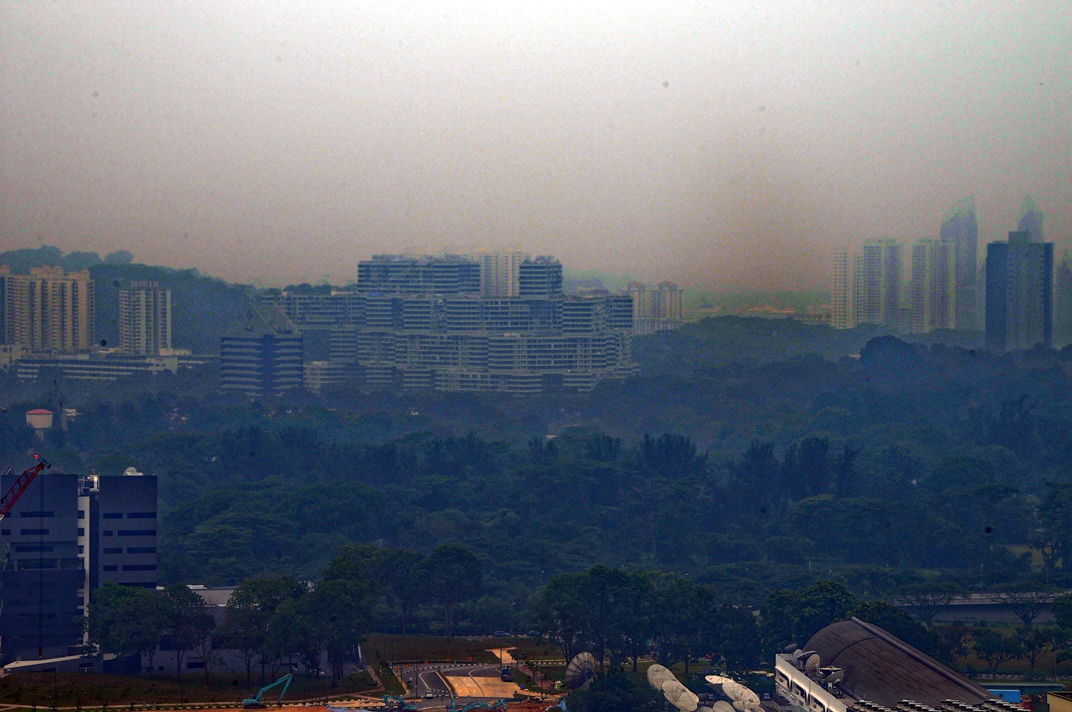}}
}
\caption{Comparison of different haze removal algorithms. From left to right, hazy images, dehazed images by DehazeNet \cite{cai}, EPDN \cite{Qu_2019_CVPR}, FFA-Net \cite{1qin2020}, MSBDN \cite{1dong2020},  DCP \cite{1he2013}, HLP \cite{CVPR16}, AOD \cite{ICCV17},  and ours,  respectively.}
\label{Fig2}
\end{figure*}

As shown in \cite{cai,Qu_2019_CVPR,1qin2020,1dong2020,ICCV17}, the data-driven algorithms perform better than the model-based algorithms on synthesized haze images. However, the performances of the data-driven algorithms are poor for real-world hazy images as indicated in  \cite{1liu2021}.   The same conclusion can be obtained from the experimental results in Fig. \ref{Fig2}. The dehazed images by  the algorithms in \cite{cai}, \cite{1qin2020}, \cite{1dong2020},  and \cite{ICCV17} are a little blurry and they are not photo-realistic. In addition, the haze is not reduced well if it is heavy. The algorithm in \cite{Qu_2019_CVPR} can be applied to overcome the problems but the generated texture  is different from real one.  Fortunately, the model-based algorithms in  \cite{1he2013}, \cite{CVPR16} and the proposed algorithm can be applied to generate  photo realistic images with the real texture. Both the algorithm in \cite{CVPR16} and the proposed algorithm preserve fine structures better than the algorithm in \cite{1he2013}. On the other hand, noise is indeed amplified in sky regions of the dehazed images by the model-based algorithms in \cite{1he2013} and \cite{CVPR16}. The experimental results validate the second inherent problem of the model-based dehazing algorithms mentioned in Section \ref{analysis}. The problem is overcome by the proposed multi-scale dehazing algorithm with the introduction of equation (\ref{air1}). Overall, the proposed algorithm outperforms other algorithms from the subjective quality point of view, especially for those hazy images with heavy haze.

Similar to \cite{1vazques2020}, the psychophysical evaluation is carried out to compare all these algorithms using the 29 real-world outdoor hazy image in the dataset \cite{data}. Twelve students are invited to select one preferred image from each set of restored images in normal lab conditions. The experimental results are shown in Table \ref{table26}. The algorithm in \cite{Li2021} and the proposed algorithm with $\rho$ as 25 are selected more frequently than other algorithms. As indicated by the numbers in the brackets $(\cdot)$, the algorithm in \cite{Li2021} and the proposed algorithm are selected more frequently for hazy images with heavy haze because they improve the visibility better than other algorithms.

\begin{table*}[htbp]
	\centering
	\caption{The number of best candidates for 29 real-world outdoor hazy images in the set \cite{data}}
	\begin{tabular}{c|c|c|c|c|c|c|c|c|c}
		\hline
DehazeNet \cite{cai} &   EPDN\cite{Qu_2019_CVPR} &	FFA-Net \cite{1qin2020} & MSBDN \cite{1dong2020} & DCP \cite{1he2013}     & HLP \cite{CVPR16} &  AOD \cite{ICCV17}   &  \cite{Li2021}  &   Ours    &   Ours    \\
   & &  &  &  &    &    &    &   $\rho=60$   &   $\rho=25$    \\		\hline
	16(4)  & 27(3) & 2 & 6 & 6 & 28  & 	 12    & 98(\textbf{20}) & 48(15) & \textbf{105}(18)\\
		\hline
		\end{tabular}
    \label{table26}
\end{table*}

\begin{table*}[htbp]
	\centering
	\caption{Average DHQI values of 79 real-world outdoor hazy images for different algorithms  (the larger, the better).}
	\begin{tabular}{c|c|c|c|c|c|c|c|c|c}
		\hline
DehazeNet \cite{cai} &   EPDN\cite{Qu_2019_CVPR} &	FFA-Net \cite{1qin2020} & MSBDN \cite{1dong2020} & DCP \cite{1he2013}     & HLP \cite{CVPR16} &  AOD \cite{ICCV17}   &  \cite{Li2021}  &   Ours    &   Ours    \\
   & &  &  &  &    &    &    &   $\rho=60$   &   $\rho=25$    \\		\hline
	 58.96  & 60.84 & 55.33 & 54.32 &51.92 &52.75  & 	 55.67    & 59.88 & 60.16 & \textbf{60.97}\\
		\hline
		\end{tabular}
    \label{table2}
\end{table*}

Besides the subjective evaluation, the quality index DHQI in \cite{1min2019} is also adopted to compare the proposed algorithm with those in \cite{cai}, \cite{Qu_2019_CVPR}, \cite{1qin2020}, \cite{1dong2020}, \cite{1he2013}, \cite{CVPR16}, \cite{ICCV17}, and \cite{Li2021}. 79 real-world outdoor hazy images are tested. Besides the 29 images in the dataset \cite{data}, 31 images are from RESIDE \cite{1haze2000}, and the remaining 19 images from \cite{1fattal2008} and the Internet. The latter 50 images are available via the link \cite{data1}. Each dehazed images of \cite{Qu_2019_CVPR} is upsampled to be consistent with the size of the input. The $L_0$ is selected as 2 in \cite{Li2021}. The average DHQI values of the 79 real-world outdoor hazy images are given in Table \ref{table2}. Clearly, the proposed algorithm also outperforms other algorithms from the objective quality point of view. The choice of $\rho$ as 25 outperforms the choice of $\rho$ as 60 from the DHQI point of view.

\begin{figure}[!htb]
\centering{
\subfigure[vehicle detection on a hazy image]{\includegraphics[width=1.6in]{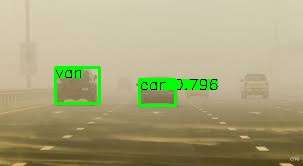}}
\subfigure[vehicle detection on a restored image]{\includegraphics[width=1.6in]{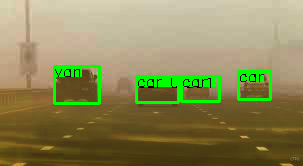}}
}
\caption{Comparison of vehicle detection on hazy and restored images. More vehicles are detected on the restored image.}
\label{Fig666888}
\end{figure}

Finally, the proposed algorithm is verified by applying it for vehicle detection \cite{vehicledetection}. As shown in Fig. \ref{Fig666888}, the detection is indeed improved for a hazy image with heavy haze.

\subsection{Limitation of the Proposed Algorithm}
\begin{figure}[!htb]
\centering{
\subfigure[one scale]{\includegraphics[width=1.6in]{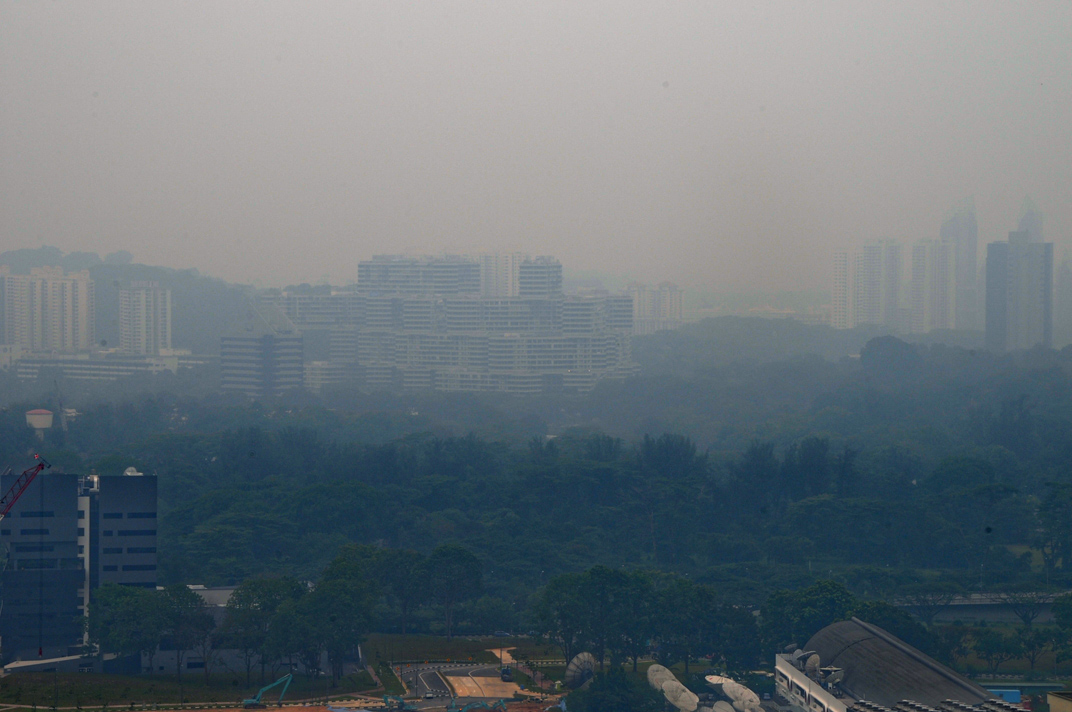}}
\subfigure[two scales]{\includegraphics[width=1.6in]{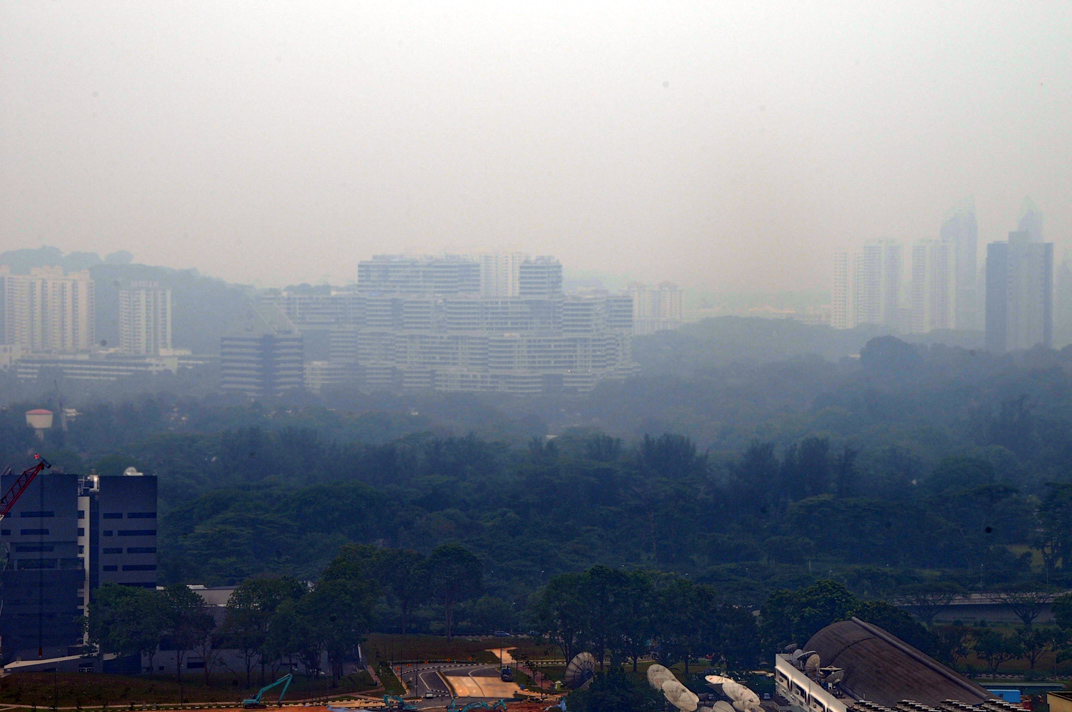}}
}
\caption{Two dehazed images by selecting the $\eta$ in the equation (\ref{eta}) as 1/4 for an image with heavy haze.}
\label{Fig666}
\end{figure}

The $\eta$ in the equation (\ref{eta}) plays an important role in the proposed dehazing algorithm. As shown in Fig. \ref{Fig2}, the haze is well removed for all hazy images if it is selected properly. However, if it is not selected properly, for example, 1/4 is selected for an image with heavy haze, the haze is not well removed as demonstrated in Fig. \ref{Fig666}(a). The proposed multi-scale dehazing algorithm can be applied to alleviate this problem a little bit as shown in Fig. \ref{Fig666}(b).

\section{conclusion Remarks}
\label{conclusion}
In this paper, the dark direct attenuation prior (DDAP) was first introduced which is applicable to all pixels in a haze image including those from sky region. A novel non-local haze line averaging was then proposed to reduce the morphological artifacts caused by the DDAP. A multi-scale hazy image model was also built up by using the Laplacian pyramid of the hazy image and the Gaussian pyramid of the transmission map. A simple multi-scale algorithm was  introduced on top of the model. Different restoration approaches were applied at different levels such that the noise is prevented from being amplified in the sky region. Experiment results illustrate that the proposed algorithm outperforms existing dehazing algorithms from both the subjective quality and the objective quality points of view.  Due to the simplicity of the proposed algorithm, it has a good potential to be adopted by mobile devices with limited computational resource.

\end{document}